\documentclass[letterpaper]{article} 
\usepackage{aaai2026}  
\usepackage{times}  
\usepackage{helvet}  
\usepackage{courier}  
\usepackage[hyphens]{url}  
\usepackage{graphicx} 
\usepackage{natbib}  
\usepackage{caption} 
\usepackage{multirow} 
\usepackage{colortbl}  
\usepackage{xcolor}  
\usepackage{booktabs}
\usepackage{subcaption}
\usepackage{makecell}
\usepackage{algorithm}
\usepackage{algorithmic}
\usepackage{bm}
\usepackage{amsmath,amsfonts}
\usepackage{newfloat}
\usepackage{listings}
\usepackage{threeparttable}
\DeclareCaptionStyle{ruled}{labelfont=normalfont,labelsep=colon,strut=off} 
\floatstyle{ruled}
\newfloat{listing}{tb}{lst}{}
\floatname{listing}{Listing}
\title{WorldRFT: Latent World Model Planning with Reinforcement Fine-Tuning for Autonomous Driving}
\author{
    Pengxuan Yang\textsuperscript{\rm 1,2,4}\equalcontrib,
    Ben Lu\textsuperscript{\rm 4}\equalcontrib,
    Zhongpu Xia\textsuperscript{\rm 1\thanks{Corresponding author.}},   
    Chao Han\textsuperscript{\rm 4},               
    Yinfeng Gao\textsuperscript{\rm 1},
    Teng Zhang\textsuperscript{\rm 4},
    Kun Zhan\textsuperscript{\rm 4},
    XianPeng Lang\textsuperscript{\rm 4},
    Yupeng Zheng\textsuperscript{\rm 1},
    Qichao Zhang\textsuperscript{\rm 1,3\thanks{Corresponding author.}}
}
\affiliations{
    \textsuperscript{\rm 1}	The State Key Laboratory of Multimodal Artificial Intelligence Systems, Institute of Automation, CAS\\
    \textsuperscript{\rm 2}	School of Advanced Interdisciplinary Sciences, UCAS\\
    \textsuperscript{\rm 3}	School of Artificial Intelligence, UCAS\\
    \textsuperscript{\rm 4}Li Auto\\

}

\usepackage{bibentry}

\begin{document}

\maketitle

\begin{abstract}

Latent World Models enhance scene representation through temporal self-supervised learning, presenting a perception annotation-free paradigm for end-to-end autonomous driving. However, the reconstruction-oriented representation learning tangles perception with planning tasks, leading to suboptimal optimization for planning.
To address this challenge, we propose WorldRFT, a planning-oriented latent world model framework that aligns scene representation learning with planning {via a hierarchical planning decomposition and local-aware interactive refinement mechanism, augmented by reinforcement learning fine-tuning (RFT) to enhance safety-critical policy performance.}
Specifically, WorldRFT integrates a vision-geometry foundation model to improve 3D spatial awareness, employs hierarchical planning task decomposition to guide representation optimization, and utilizes local-aware iterative refinement to derive a planning-oriented driving policy. Furthermore, we introduce Group Relative Policy Optimization (GRPO), which applies trajectory Gaussianization and collision-aware rewards to fine-tune the driving policy, yielding systematic improvements in safety. WorldRFT achieves state-of-the-art (SOTA) performance on both open-loop nuScenes and closed-loop NavSim benchmarks. On nuScenes, it reduces collision rates by 83\% (0.30\% → 0.05\%). On NavSim, using camera-only sensors input, it attains competitive performance with the LiDAR-based SOTA method DiffusionDrive (87.8 vs. 88.1 PDMS). 

\end{abstract}

\begin{links}
    \link{Code}{https://github.com/pengxuanyang/WorldRFT}
\end{links}

\begin{figure}
    \centering
    \includegraphics[width=\linewidth]{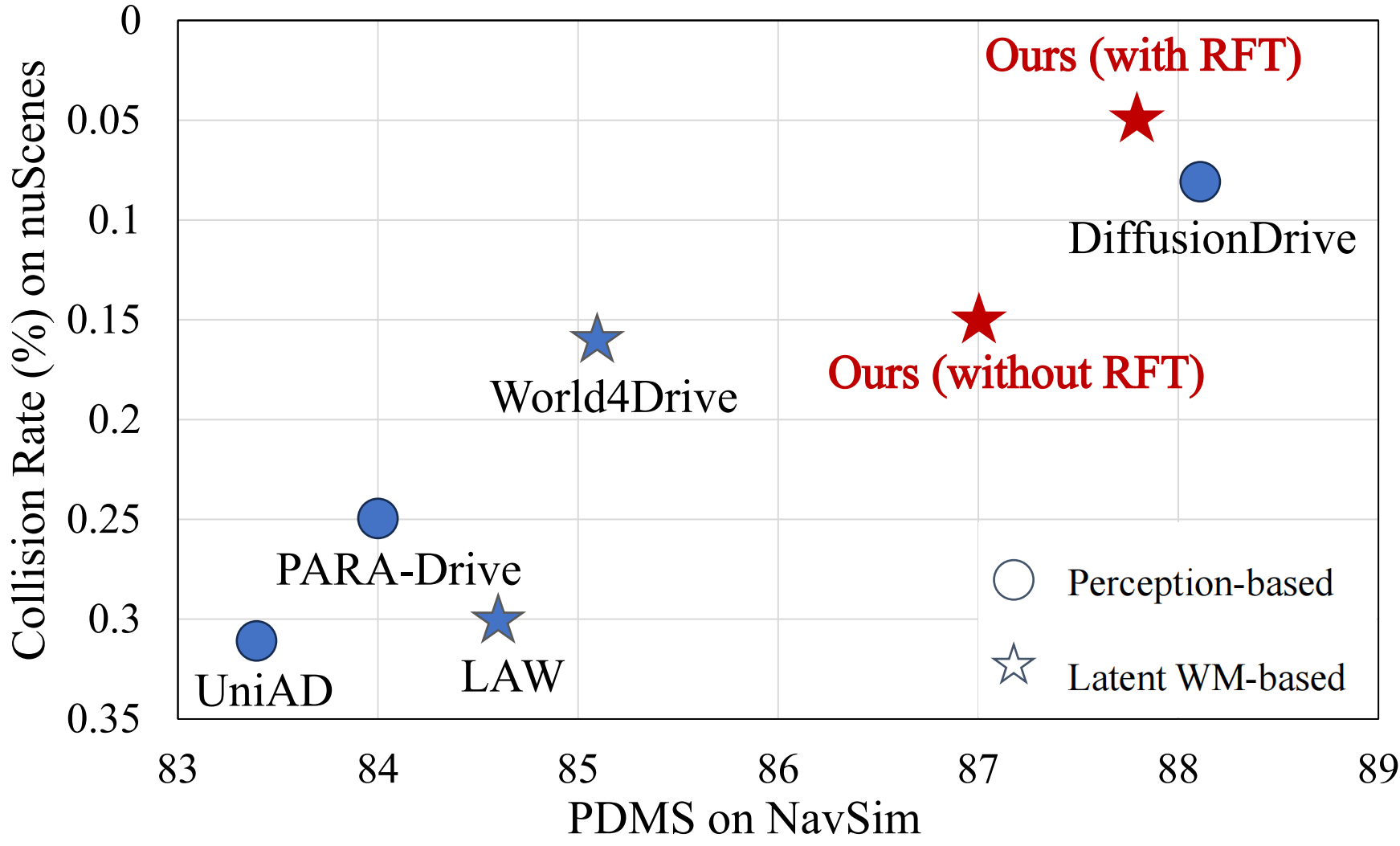}
    \caption{Performance comparison of SOTA methods on open-loop nuScenes and closed-loop NavSim benchmarks.}
    \label{fig:fig1}
\end{figure}

\section{Introduction}
Traditional end-to-end autonomous driving methods \cite{chen2024vadv2,jia2025drivetransformer,jiang2023vad,sun2024sparsedrive} often rely on multi-task architectures incorporating auxiliary perception modules such as object detection, online mapping, and occupancy prediction to improve scene understanding. Recently, a new paradigm of constructing unified latent world model (WM) representations from raw surround-view images \cite{zheng2025world4drive,li2024enhancing} has enabled slight end-to-end autonomous driving through temporal self-supervised learning.

Despite this progress, existing latent WM suffers from a critical misalignment between reconstruction-oriented representations and planning requirements, manifesting in three key challenges:
(1) Lacking spatial awareness: current reconstruction objectives produce generic representations with limited 3D spatial awareness, which is crucial for trajectory planning. 
Although World4Drive \cite{zheng2025world4drive} attempts to incorporate explicit 3D information (depth map) through monocular depth estimation models \cite{hu2024metric3d}, this approach suffers from cross-view inconsistencies, limiting the model's global comprehension of complex driving scenarios. 
(2) Inefficient planning interaction mechanisms: existing models employ a single global planning query to generate trajectories from entire feature maps, inadequately capturing local structures essential for fine-grained planning. This results in dispersed attention and inability to effectively comprehend and extract planning-relevant scene representations,  as shown in Figure~\ref{fig:4}, leading to suboptimal feature utilization for planning decisions.
(3) Limited safety awareness: perception-based imitation learning approaches, such as PARA-Drive \cite{weng2024para}, DiffusionDrive\cite{liao2024diffusiondrive}, focus on minimizing trajectory deviations from expert demonstrations without incorporating explicit safety objectives.
This paradigm lacks active collision avoidance, merely treating all deviations from expert trajectories equally without understanding underlying safety principles or proactively identifying potential risks in novel scenarios.

To address these challenges, we propose \textbf{WorldRFT}, a planning-oriented latent world model framework that aligns representation learning with planning requirements. 
First, addressing the weak 3D spatial understanding, we design a \textbf{Spatial-aware World Encoder (SWE)} that constructs 3D spatially-aware representations by incorporating the visual-geometric foundation model VGGT~\cite{wang2025vggt}, establishing a robust spatial understanding for planning tasks. 
Second, resolving the inefficiency of planning interaction mechanisms, we propose a \textbf{Hierarchical Planning Refinement (HPR)} module that decomposes complex end-to-end planning into three parallel subtasks which describe planning in hierarchical dimension: target region localization, spatial path planning, and temporal trajectory prediction. Building upon this decomposition, we develop a local-aware iterative refinement module that dynamically samples and fuses task-relevant local features from the latent space, ensuring both global consistency and local precision. 
Third, to enhance safety-critical planning capabilities, we design a safety-aware \textbf{Reinforcement learning Fine-Tuning (RFT)} phase based on Group Relative Policy Optimization (GRPO) \cite{shao2024grpo} to optimize explicit safety objectives. By gaussianizing predicted trajectories and employing collision-aware rewards, our approach transitions from passive behavior cloning to active collision avoidance.

Experiments demonstrate that WorldRFT achieves state-of-the-art (SOTA) performance on both open-loop nuScenes and closed-loop NavSim benchmarks, as shown in Figure~\ref{fig:fig1}. On the nuScenes dataset, compared to the baseline LAW \cite{li2024enhancing}, average displacement error decreases by \textbf{21\% }($0.61\text{m} \rightarrow 0.48\text{m}$) and collision rate drops by \textbf{83\%} ($0.30\% \rightarrow 0.05\%$). On the NavSim dataset, the PDMS metric improves by 3.2 points ($84.6 \rightarrow \textbf{87.8}$), achieving the best performance among vision-only solutions and approaching the LiDAR-based SOTA method DiffusionDrive (88.1).

Our main contributions are summarized as follows:
\begin{itemize}
    \item We propose the first planning-oriented latent world modeling paradigm that deeply aligns representation learning with planning tasks through the spatial geometric prior fusion and hierarchical planning interaction and local-aware refinement.
    \item We introduce a reinforcement fine-tuning phase that enhances safety planning via explicit safety rewards and GRPO method, transitioning from behavioral imitation to proactive collision avoidance.
    \item Our method achieves SOTA performance on both nuScenes and NavSim benchmarks, with substantial improvements in critical safety metrics.
\end{itemize}

\section{Related Works}
\subsection{End-to-End Autonomous Driving}

End-to-end autonomous driving models utilize the  imitation learning framework with perception
annotations to directly map sensor inputs to trajectory outputs\cite{10547289,yang2025uncad}. For instance, UniAD and VAD \cite{hu2023planning, jiang2023vad} fuse multiple perception and planning tasks within a cascaded BEV-based architecture, while PARA-Drive and ONE-Drive \cite{weng2024para,zheng2024preliminary} leverage parallel architecture to enhance efficiency. 
Methods like SparseDrive~\cite{sun2024sparsedrive} integrate detection, tracking, and mapping through symmetric sparse perception modules. 
Despite their advantages, these approaches require high-precision 3D perception annotations, leading to higher development complexity and costs.

\subsection{World Models for Autonomous Driving}

World models predict future scene states to improve dynamic environment understanding and support safer path planning. For instance, GAIA-1 \cite{hu2023gaia} creates realistic driving scenarios from multi-modal inputs and allows detailed control over vehicle behavior, while DriveDreamer4D \cite{zhao2025drivedreamer4D} generates novel trajectory videos from real driving data to enhance 4D reconstruction. Methods like DriveWorld and PreWorld~\cite{min2024driveworld, li2025semi, gu2024dome, jin2025occtens} generate future occupancy and flow fields from BEV embeddings. 
In contrast, self-supervised approaches like LAW \cite{li2024enhancing} and SSR \cite{li2025navigationguided} model future scene dynamics without labeled data, but their relatively coarse planning frameworks can limit planning effectiveness, indicating a need for more sophisticated architectural designs.

\subsection{Reinforcement Learning for Autonomous Driving}
Reinforcement learning (RL) is increasingly important in autonomous driving for complex decision-making. 
RAD \cite{gao2025rad} builds virtual environments with 3D Gaussian Splatting to facilitate exploration and policy learning in diverse scenarios. 
AlphaDrive \cite{jiang2025alphadrive} introduces RL to enhance planning and training of vision-language models. 
Given RL’s advantage over imitation learning in handling causal confounding, we adopt RL for trajectory planning to improve safety and reduce collision risk in autonomous driving.
\section{Methodology}
\subsection{Overview}

\begin{figure*} 
    \centering
    \includegraphics[width = \linewidth]{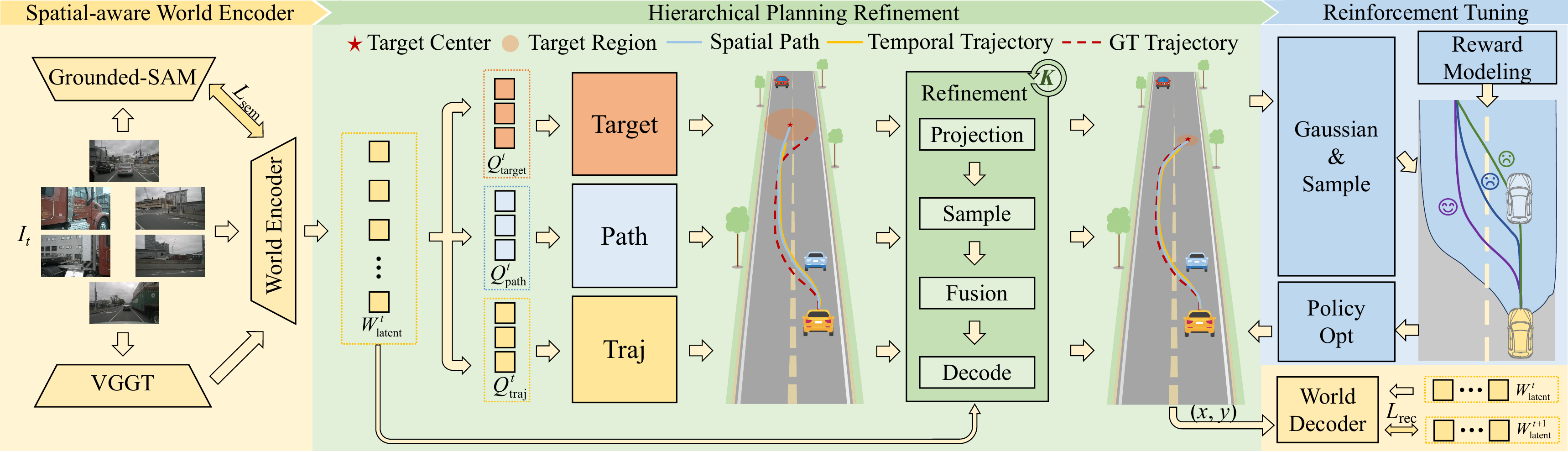}
    \caption{WorldRFT consists of three key modules: 1) Spatial-aware World Encoder (yellow) that extracts geometry-rich latent world representations from RGB images; 2) Hierarchical Planning Refinement Module (green) that efficiently captures critical information highly correlated with driving decisions through a refined planning task design; 3) Safety-aware Reinforcement Learning Fine-tuning phase (blue) that generates safer planning outcomes via reinforcement learning optimization. These modules work synergistically to deliver high-quality end-to-end autonomous driving planning.}
    \label{fig:main}
\end{figure*}

As illustrated in Figure~\ref{fig:main}, WorldRFT comprises three key modules that form a complete pipeline of ``scene understanding $\rightarrow$ planning decisions $\rightarrow$ safety optimization'': (1) SWE constructs spatial-rich latent world representations from RGB images; (2) HPR efficiently extracts critical information strongly correlated with driving decisions; (3) RFT generates safer planning results through reinforcement learning optimization. These modules work synergistically to achieve safer end-to-end autonomous driving planning.

\subsection{Spatial-aware World Encoder}

The spatial-aware world encoder constructs 3D spatially-aware latent world representations serving planning tasks from surround-view inputs through the geometric foundation model VGGT~\cite{wang2025vggt}.

\textbf{Basic Visual Feature Encoding.}
Given multi-view images $I_t \in \mathbb{R}^{M\times H \times W \times 3}$ at time $t$, we first extract features $F_t \in \mathbb{R}^{M\times h\times w\times D}$ through the image backbone. Following World4Drive~\cite{zheng2025world4drive}, we then utilize the vision-language foundation model Grounded-SAM for semantic supervised learning through the loss function $\mathcal{L}_{\text{sem}}$.

\textbf{3D Spatial Encoding.}
Spatial understanding capability is crucial for planning tasks. VGGT, benefiting from its diverse training data and elegant feed-forward architecture, provides multi-view consistent geometric-rich prior from 2D surround images $I_t$. We leveraging VGGT's structural priors to enhance spatial awareness. Specifically, to maintain generic 3D aware capacity, we employ a frozen VGGT as a 3D spatial encoder and extract 3D tokens from its final layer:
\begin{equation}
\varepsilon_l(I_t) = \{t_c, t_r, t_{3D}\}
\end{equation}
where $l$ denotes the layer index, and $t_c$, $t_r$, and $t_{3D}$ represent camera, register, and 3D tokens, respectively. 
We subsequently incorporate the 3D tokens $t_{3D}$ into the image features $F_t$ to construct spatial-aware latent world representations. Specifically, we propose a lightweight fusion module that integrates ResNet and VGGT features via a single cross-attention layer, where the 2D visual features $F_t$ act as queries and the VGGT-derived 3D features $t_{3D}$ are used as keys and values. This process ultimately yields the unified latent space visual representation $W^{t}_{\text{latent}}$, enabling effective spatial understanding without requiring explicit 3D inputs:
\begin{equation}
W^{t}_{\text{latent}} = \text{C-A}(F_t, t_{3D})
\end{equation}
where $\text{C-A}$ denotes Cross-Attention layers.

\subsection{Hierarchical Planning Refinement}
\label{sec:extraction}
In this section, we decompose end-to-end planning into three parallel subtasks that extract hierarchical features with distinct supervision: target region localization, spatial path planning, and temporal trajectory prediction. These tasks capture complementary aspects of planning and are unified through a query interaction framework and refined via local-aware iterative refinement.

\textbf{Unified Query Interaction Framework.}
To facilitate coordination among parallel subtasks while maintaining their unique functionalities, we design a unified query interaction framework. Specifically, each subtask is initialized with dedicated queries: $Q_{\text{target}}$, $Q_{\text{path}}$, and $Q_{\text{traj}}$, which then independently aggregate hierarchical planning-relevant features from the latent world representation:
\begin{equation}
[Q'_{\text{target}}, Q'_{\text{path}}, Q'_{\text{traj}}] = \text{C-A}([Q_{\text{target}}, Q_{\text{path}}, Q_{\text{traj}}], W^{t}_{\text{latent}})
\end{equation}
To facilitate inter-task communication, these queries undergo concatenation and self-attention, enabling mutual awareness of planning intentions:
\begin{equation}
[Q''_{\text{target}}, Q''_{\text{path}}, Q''_{\text{traj}}] = \text{S-A}(\text{Concat}[Q'_{\text{target}}, Q'_{\text{path}}, Q'_{\text{traj}}])
\end{equation}

\textbf{Target Region Localization.}
Predicting a deterministic target point for navigation~\cite{xing2025goalflow} is an ill-posed problem due to the inherent uncertainty of valid target points in planning scenarios. Therefore, 
rather than predicting deterministic target points, we model targets as probabilistic regions using Laplace distributions. In detail, we parameterize the target region as:
\begin{equation}
(\mu, b) = \text{MLP}(Q''_{\text{target}})
\end{equation}
where $\mu \in \mathbb{R}^2$ denotes the region center and $b \in \mathbb{R}^2$ represents the scale parameters, quantifying the spatial extent of the target region. This formulation enables the model to learn target region distributions rather than fitting to a single deterministic point.
Moreover, the probabilistic representation naturally captures scene complexity, thereby providing adaptive information that can be leveraged by the iterative refinement module to guide feature fusion.

The probabilistic representation is trained using the negative log-likelihood loss:
\begin{equation}\label{nll}
\mathcal{L}_{\text{Laplace-NLL}} = \log(2b) + \frac{\|\mathbf{y} - \boldsymbol{\mu}\|_1}{b}
\end{equation}

Here, the scale parameter $b$ naturally reflects scene complexity, with larger values corresponding to more challenging scenarios that demand cautious planning. As a conditioning signal, $b$ modulates feature fusion within the uncertainty-aware local adaptation module, thereby enabling adaptive planning in accordance with scene uncertainty.

\textbf{Spatial Path Planning.}
This module generates a spatial path connecting the current position to the target region. To ensure spatial consistency, we construct the spatial path ground truth by uniformly sampling future trajectory points at fixed spatial intervals, rather than temporal intervals ~\cite{li2024enhancing, li2025navigationguided, zheng2025world4drive, jiang2023vad}. Furthermore, we decode $N$ spatial path points through an MLP network:
\begin{equation}
T_{\text{path}} = \text{MLP}(Q''_{\text{path}})
\end{equation}
where $T_{\text{path}} \in \mathbb{R}^{N \times 2}$ denotes path points sampled at 2-meter intervals, with each point denoting coordinates $(x, y)$.

\textbf{Temporal Trajectory Prediction.}
In this module, we decode future temporal trajectories for $T$ timesteps through a trajectory decoder:
\begin{equation}
T_{\text{traj}} = \text{MLP}(Q''_{\text{traj}})
\end{equation}
where $T_{\text{traj}} \in \mathbb{R}^{T \times 2}$ represents trajectory points at fixed temporal intervals (0.5 seconds), with each point containing coordinates $(x, y)$.

\begin{figure}
    \centering
    \includegraphics[width=\linewidth]{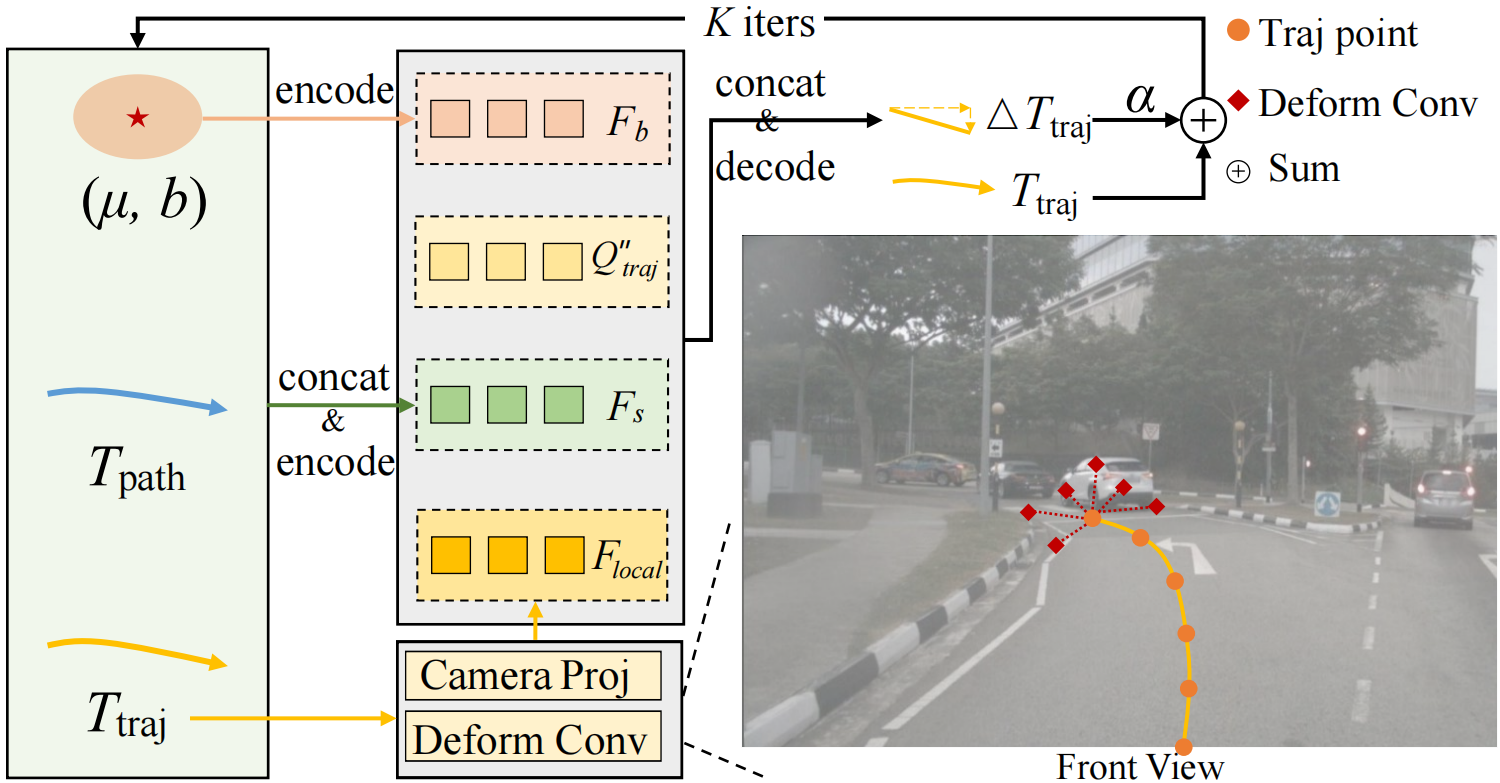}
    \caption{Architecture of the Local-aware Iterative Refinement module. The module refines preliminary planning results $(\mu,b), T_{\text{path}}, T_{\text{traj}}$ through $K$ iterations. Using $T_{\text{traj}}$ as an example, each iteration: (1) encodes global planning states, (2) projects trajectory points via camera parameters, (3) samples local features using deformable convolution, and (4) fuses them with global information and uncertainty representation. Residual connections enable incremental updates for adaptive local adjustment.}
    \label{fig:subfig2}
\end{figure}

\textbf{Local-aware Iterative Refinement.}
This module iteratively refines the initial planning outputs by integrating local scene information. It takes the preliminary results $(\mu, b)$, $T_{\text{path}}$, and $T_{\text{traj}}$ as inputs, and refines the outputs of all three subtasks over $K$ iterations. Using temporal trajectory refinement as an example, each iteration involves the following key steps as shown in Figure~\ref{fig:subfig2}:

First, we encode current planning information $(\mu, b)$, $T_{\text{path}}$, and $T_{\text{traj}}$ into a unified state representation $F_s$:
\begin{equation}
F_s = \text{MLP}(\text{Concat}[(\mu^{(k)}, b^{(k)}), T^{(k)}_{\text{path}}, T^{(k)}_{\text{traj}}])
\end{equation}

Subsequently, the trajectory points are projected onto the latent space feature map using the camera parameters, yielding corresponding positions $P_{\text{proj}}$. Deformable convolution is then applied to adaptively sample the local features $F_{\text{local}}$ at these positions:
\begin{equation}
P_{\text{proj}} = \text{CameraProject}(T^{(k)}_{\text{traj}})
\end{equation}
\begin{equation}
F_{\text{local}} = \text{DeformConv}(W^{t}_{\text{latent}}, P_{\text{proj}})
\end{equation}

Next, local and global features are fused, guided by the scale parameter $b$ obtained from target region modeling, which acts as a conditioning signal:
\begin{equation}
F_b = \text{MLP}(b^{(k)})
\end{equation}
\begin{equation}
F_{\text{fusion}} = \text{MLP}(\text{Concat}[F_{\text{local}}, Q''_{\text{traj}}, F_s, F_b])
\end{equation}
where $Q''_{\text{traj}}$ represents the global planning intention.

Finally, independent prediction heads generate the corresponding trajectory offsets $\Delta T_{\text{traj}}^{(k)}$, which are then used to incrementally update the trajectory via residual connections:
\begin{equation}
\Delta T_{\text{traj}}^{(k)} = \text{MLP}(F_{\text{fusion}})
\end{equation}
\begin{equation}
T^{(k+1)}_{\text{traj}} = T^{(k)}_{\text{traj}} + \alpha \Delta T^{(k)}_{\text{traj}}
\end{equation}
where $\alpha = 0.1$ is the update step size. 

\textbf{Training Loss.}
We pretrain our model guided by the latent world modeling, which enables self-supervised learning without requiring perception annotations by leveraging temporal consistency in future world prediction.
Following \cite{li2024enhancing}, we use a world decoder to predict future latent world representations $\widehat{W}^{t+1}_{\text{latent}}$ from the current latent representation $ W^{t}_{\text{latent}}$ and the temporal trajectory $T_{\text{traj}}$.
We self-supervise world prediction through the MSE loss:
\begin{equation}\label{mse}
\mathcal{L}_{\text{rec}} = \text{MSE}(\widehat{W}^{t+1}_{\text{latent}}, W^{t+1}_{\text{latent}})
\end{equation}

Additionally, we employ L1 loss $\mathcal{L}_{\text{traj}}$ to align both the spatial path $T_{\text{path}}$ and temporal trajectory $T_{\text{traj}}$ with expert demonstrations. The final training objective combines multiple loss terms:

\begin{equation}
\mathcal{L}_{\text{Total}} = \alpha\mathcal{L}_{\text{sem}} + \beta\mathcal{L}_{\text{rec}} + \gamma\mathcal{L}_{\text{target}} + \eta\mathcal{L}_{\text{traj}}
\end{equation}
where $\alpha, \beta, \gamma$ and $\eta$ are hyper-parameters.

\subsection{Reinforcement Learning Fine-tuning}
\label{sec:finetuning}
In this phase, the module introduces collision-aware rewards, extends trajectories to probability distributions for exploration, and applies GRPO to improve safety while maintaining planning accuracy.

\textbf{Collision-aware Reward Design.}
Our RL framework targets collision risks between the ego vehicle and surrounding agents. The collision-aware reward is based on the distance between the ego vehicle’s bounding box and those of nearby agents along the trajectory: negative distances (collisions) incur penalties, while non-collisions yield zero reward:
\begin{equation} \label{eq:4-n}
    r = 
    \begin{cases}
        -1 & \text{if collision happens} \\
         0 & \text{if no collision happens}
    \end{cases}
\end{equation}

\textbf{Gaussianized-Trajectory Modeling.}
Since the above trajectory prediction is formulated as a regression task, effective application of RL requires recasting it as a probabilistic, distribution-based problem. To this end, we model trajectories using Gaussian distributions, where the predicted trajectory serve as the means ${\bm{\mu }_\theta }$, and an auxiliary variance network adaptively estimates the trajectory variances ${{\bm{\Sigma }}_\theta }$.

\textbf{Policy Optimization.}
Based on the world model structure, we optimize generated trajectories using GRPO \cite{shao2024grpo}. Following the GRPO framework, we estimate advantage functions based on relative rewards within sample groups. Specifically, a group of $G$ trajectories is sampled using the previously described latent world model, which serves as the policy ${\pi _{\theta}}$: $\left\{ {T_{\text{traj}_0},T_{\text{traj}_1},...,T_{\text{traj}_{G-1}}} \right\}$. For each point along these trajectories, the relative reward is defined as the normalized per-point reward value:
	\begin{equation} \label{eq:4-3}
\tilde r_j^{T_{\text{traj}_i}} = \frac{{r_j^{T_{\text{traj}_i}} - {\rm{mean}}\left( {{\bm{r}_j}} \right)}}{{{\rm{std}}\left( {{\bm{r}_j}} \right)}}
\end{equation}
where $r_j^{T_{\text{traj}_i}}$ denotes the value of the original reward function, and ${\bm{r}_j} = \left\{ {r_j^{T_{\text{traj}_0}},r_j^{T_{\text{traj}_1}},...,r_j^{T_{\text{traj}_{G-1}}}} \right\}$ denotes the values of the group-based reward function.

Given that the predicted trajectories are represented as time-differential increments, each increment at the current time step influences both the current and all subsequent points, leading to potential error accumulation. Accordingly, the relative advantage function for the $j$-th trajectory point is defined as:
	\begin{equation} \label{eq:4-4}
	Adv_j^{T_{\text{traj}_i}} = \sum\limits_{t \ge j} {\tilde r_t^{T_{\text{traj}_i}}} 
\end{equation}

The policy optimization objective function becomes:
\begin{equation} \label{eq:4-27}
{J}\left( \theta  \right) = \frac{1}{G}\sum\limits_i {\sum\limits_j {\left\{ {\min \left[ {{f_1},{f_2}} \right] - \beta {D_{KL}}\left( {{\pi _\theta }||{\pi _{{\theta _{ref}}}}} \right)} \right\}} }
\end{equation}
where:
\begin{equation} \label{eq:4-28}
{f_1} = \left( {\frac{{{\pi _\theta }}}{{{\pi _{\theta old}}}}} \right)Adv_j^{T_{\text{traj}_i}}
\end{equation}

\begin{equation} \label{eq:4-29}
{f_2} = \text{clip}\left( {\frac{{{\pi _\theta }}}{{{\pi _{{\theta _{old}}}}}},1 - \varepsilon ,1 + \varepsilon } \right)Adv_j^{T_{\text{traj}_i}}
\end{equation}
where $\varepsilon$ limits the update step size of the policy function to ensure algorithmic stability, and $\beta$ is the coefficient for the Kullback-Leibler (KL) divergence. ${\pi _{{\theta _{old}}}}$ denotes the old policy, and ${\pi _{{\theta _{ref}}}}$ represents the reference policy, which is the pre-trained model. As mentioned above, we model the output of the trajectory as a Gaussian distribution with mean vector ${\bm{\mu }_\theta }$ and covariance matrix ${{\bm{\Sigma }}_\theta }$. The reference model provides a deterministic result ${\bm{\mu }_{ref}} = T_{\text{traj}_{ref}}$, which is used to compute the negative log-likelihood (NLL) loss under the Gaussian distribution. Thus, the KL divergence is defined as:

\begin{equation} \label{eq:4-30}
{D_{KL}} = \frac{1}{2}\left[ \begin{array}{l}
	\log |{{\bf{\Sigma }}_\theta }| + {({\bm{\mu }_{ref}} - {\bm{\mu }_\theta })^T}{\bf{\Sigma }}_{{\rm{\theta}}}^{ - 1}({\bm{\mu }_{ref}} - {\bm{\mu }_\theta })\\
	+ 2\log (2\pi )
\end{array} \right]
\end{equation}

\begin{table*}[htb]
    \centering
    \small
    \begin{threeparttable}
    \resizebox{\textwidth}{!}{ 
    \begin{tabular}{l|c|ccc>{\columncolor[gray]{0.9}}c|ccc>{\columncolor[gray]{0.9}}c}
        \toprule
        \multirow{2}{*}{\text{Method}} & \multirow{2}{*}{\text{Training Approach}} & \multicolumn{4}{c|}{L2 (m) $\downarrow$} & \multicolumn{4}{c}{\text{Collision Rate (\%) $\downarrow$}}  \\
        & &\text{1s} & \text{2s} & \text{3s} & \text{Avg.} & \text{1s} & \text{2s} & \text{3s} & \text{Avg.} \\
        \midrule
        ST-P3~\cite{hu2022st} & P-IL & 1.33 & 2.11 & 2.90 & 2.11 & 0.23 & 0.62 & 1.27 & 0.71  \\
        OccNet~\cite{tong2023scene} & P-IL & 1.29 & 2.13 & 2.99 & 2.13 & 0.21 & 0.59 & 1.37 & 0.72 \\
        UniAD~\cite{hu2023planning} & P-IL & 0.48 & 0.96 & 1.65 & 1.03 & {0.05} & 0.17 & 0.71 & 0.31  \\
        VAD~\cite{jiang2023vad} & P-IL & 0.41 & 0.70 & 1.05 & 0.72 & 0.07 & 0.18 & 0.43 & 0.23  \\
        UncAD~\cite{yang2025uncad} & P-IL & 0.33 & 0.59 & 0.94 & 0.62 & 0.10 & 0.14 & 0.28 & 0.17  \\
        PPAD~\cite{chen2024ppad} & P-IL & 0.31 & 0.56 & 0.87 & 0.58 & 0.08 & {0.12} & 0.38 & 0.19 \\
        PARA-Drive~\cite{weng2024para} & P-IL & 0.25 & 0.46 & \textbf{0.74} & 0.48 & 0.14 & 0.23 & 0.39 & 0.25  \\
        GenAD~\cite{zheng2024genad} & P-IL & 0.28 & 0.49 & 0.78 & 0.52 & 0.08 & 0.14 & 0.34 & 0.19 \\
        LAW~\cite{li2024enhancing} (Perception-based) & P-IL & {0.24} & 0.46 & 0.76 & 0.49 & 0.08 & 0.10 & 0.39 & 0.19  \\
        DiffusionDrive~\cite{liao2024diffusiondrive} & P-IL & {0.27} & 0.54 & 0.90 & 0.57 & 0.03 & 0.05 & \textbf{0.16} & 0.08  \\
        \midrule
        Epona~\cite{zhang2025epona} & SS-L & 0.61 & 1.17 & 1.98 & 1.25 & 0.01 & 0.22 & 0.85 & 0.36 \\
        LAW~\cite{li2024enhancing} (Perception-free) & SS-L & 0.26 & 0.57 & 1.01 & 0.61 & 0.14 & 0.21 & 0.54 & 0.30 \\
        World4Drive~\cite{zheng2025world4drive} & SS-L & 0.23& 0.47& 0.81& 0.50& 0.02 & 0.12 & 0.33 &0.16  \\
        Ours (without RFT) & SS-L & \textbf{0.21}& \textbf{0.44}& {0.76}& \textbf{0.47}& {0.10}&{0.11}& {0.23}& {0.15}\\
        Ours (with RFT) & SS-L \& RL & {0.22} & \textbf{0.44}& {0.77}& {0.48}& \textbf{0.00}&\textbf{0.00}& \textbf{0.16}& \textbf{0.05}\\
        \midrule
        \rowcolor[gray]{0.97} \color{gray}SSR\dag~\cite{li2025navigationguided} & \color{gray}SS-L & \color{gray}0.19 & \color{gray}0.36 & \color{gray}0.62 & \color{gray}0.39 & \color{gray}0.00 & \color{gray}0.10 & \color{gray}0.20 & \color{gray}0.13 \\
        \rowcolor[gray]{0.97} \color{gray}Ours (with RFT)\dag & \color{gray}SS-L \& RL & \color{gray}0.15 & \color{gray}0.30 & \color{gray}0.56 & \color{gray}0.33 & \color{gray}0.00 & \color{gray}0.00 & \color{gray}0.12 & \color{gray}0.04 \\
        \bottomrule
    
    \end{tabular}
    }
    \begin{tablenotes}[left]
    \item P-IL: Perception-based Imitation Learning; SS-L: Self-Supervised Learning; RL: Reinforcement Learning.
    \item \dag\ Represents methods incorporating ego status.
    \end{tablenotes}
    \end{threeparttable}
    \caption{End-to-end planning results on nuScenes benchmark~\cite{caesar2020nuscenes}}
    \label{tab1}
\end{table*}

\begin{table*}[!ht]
    \centering
    \small
    \begin{threeparttable}
    \resizebox{\textwidth}{!}{ 
\begin{tabular}{l|cc|cccccccc}
    \toprule
    Method & Training Approach & Input & NC $\uparrow$ &DAC $\uparrow$ & TTC $\uparrow$& Comf. $\uparrow$ & EP $\uparrow$ & \cellcolor{gray!30}PDMS $\uparrow$  \\
    \midrule
    UniAD~\cite{hu2023planning} & P-IL & C & 97.8 & 91.9 & 92.9 & \textbf{100.0} & 78.8 & \cellcolor{gray!30}83.4 \\
    PARA-Drive~\cite{weng2024para} & P-IL & C & \underline{97.9} & 92.4 & 93.0 & 99.8 & 79.3 & \cellcolor{gray!30}84.0 \\
    LTF~\cite{prakash2021multi} & P-IL & C & 97.4 & 92.8 & 92.4 & \textbf{100.0} & 79.0 & \cellcolor{gray!30}83.8 \\
    Transfuser~\cite{prakash2021multi} & P-IL & C \& L & 97.7 & 92.8 & 92.8 & \textbf{100.0} & 79.2 & \cellcolor{gray!30}84.0 \\
    VADv2~\cite{chen2024vadv2} & P-IL & C \& L & 97.2 & 89.1 & 91.6 & \textbf{100.0} & 76.0 & \cellcolor{gray!30}80.9 \\
    Hydra-MDP~\cite{li2024hydra} & P-IL & C \& L & \underline{97.9} & 91.7 & 92.9 & \textbf{100.0} & 77.6 & \cellcolor{gray!30}83.0 \\
    DiffusionDrive~\cite{liao2024diffusiondrive} & P-IL & C \& L & \textbf{98.2}  & \underline{96.2}  & \textbf{94.7}  & \textbf{100.0}  & \textbf{82.2}  & \cellcolor{gray!30}\textbf{88.1}\\
    \midrule
    Epona \cite{zhang2025epona} & SS-L & C & {97.9}  & 95.1  & {93.8}  & \textbf{99.9}  & 80.4  & \cellcolor{gray!30}86.2\\
    LAW (Perception-free)~\cite{li2024enhancing} & SS-L & C & 96.4  & 95.4  & 88.7  & 99.9  & \underline{81.7}  & \cellcolor{gray!30}{84.6}\\
    DriveX \cite{shi2025drivex} & SS-L & C & {97.5}  & 94.0  & {93.0}  & \textbf{100.0}  & 79.7  & \cellcolor{gray!30}84.5\\
    World4Drive \cite{zheng2025world4drive} & SS-L & C & {97.4}  & 94.3  & {92.8}  & \textbf{100.0}  & 79.9  & \cellcolor{gray!30}85.1\\
    Ours (without RFT) & SS-L & C & {97.5}  & {96.0}  & \underline{94.0}  & \textbf{100.0}  & {80.9}  & \cellcolor{gray!30}{87.0}\\
    Ours (with RFT) & SS-L \& RL & C & {97.8}  & \textbf{96.8}  & \underline{94.0}  & \textbf{100.0}  & \underline{81.7}  & \cellcolor{gray!30}\underline{87.8}\\
    \bottomrule
\end{tabular}%
}
    \begin{tablenotes}[left]
    \item P-IL: Perception-based Imitation Learning; SS-L: Self-Supervised Learning; RL: Reinforcement Learning.
    \item C: Camera modality; L: LiDAR modality.
    \end{tablenotes}
    \end{threeparttable}
    \caption{End-to-end planning results on NavSim benchmark~\cite{im2024navsim}}
    \label{tab2}
\end{table*}

\begin{figure*}[!t]
    \centering
    \includegraphics[width=1\linewidth]{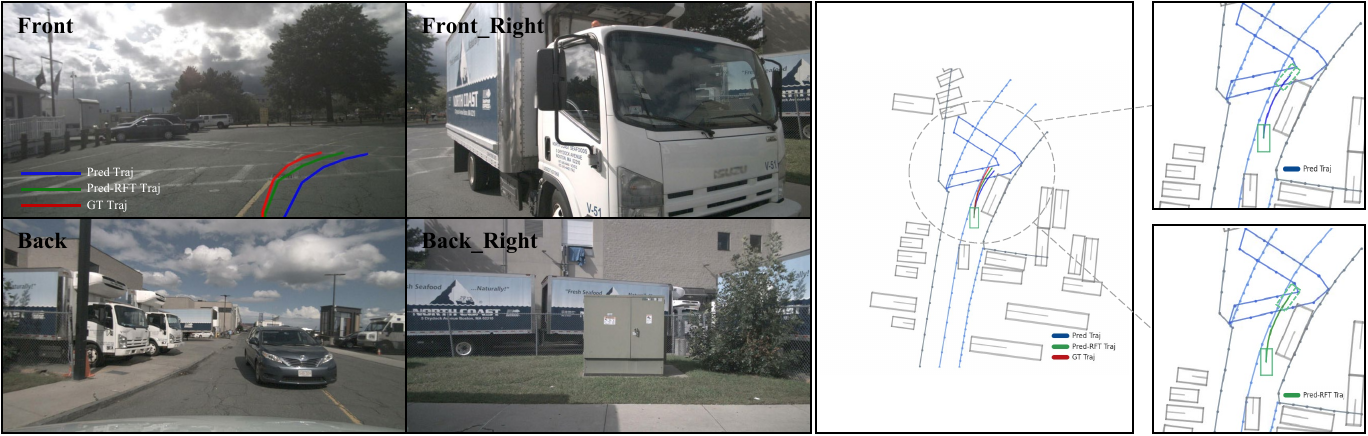}
    \caption{Visualization of planning trajectories, where the red line is the ground truth, the blue line is the pretrained-only trajectory, and the green line shows the RFT-trajectory. Obviously, the pre-trained trajectory dangerously approaches obstacles, risking collisions. In contrast, the RFT-trajectory proactively adjusts to maintain safe distances, avoiding collisions.}
    \label{fig:3}
\end{figure*}

\textbf{RFT Training Loss.}
We fine-tune the model using the GRPO objective function and KL loss. Therefore, the loss for reinforcement learning can be written as:

\begin{equation}
\mathcal{L}_{\text{RL}} = {-J}\left( \theta  \right) + \lambda{D_{KL}}
\end{equation}

\section{Experiment}

\subsection{Overview}
This section presents the experimental results of the proposed method, including ablation studies for each component. 
For completeness, detailed descriptions of the dataset, experimental setup, evaluation metrics, hyperparameter settings, scalability analysis, and additional visualizations are provided in the supplementary material.
\subsection{Benchmark}

We comprehensively evaluate our method on the open-loop nuScenes benchmark \cite{caesar2020nuscenes} and the closed-loop NavSim benchmark \cite{im2024navsim}. On nuScenes, we report displacement error (L2) and collision rate (CR) for trajectory prediction. For NavSim, we adopt the PDM Score (PDMS), which combines performance across five dimensions: no at-fault collision (NC), drivable area compliance (DAC), time-to-collision (TTC), ride comfort (Comf.), and ego progress (EP).

\subsection{Main Results}

As demonstrated in Table~\ref{tab1}, we compare our framework with several SOTA methods. Specifically, WorldRFT achieves SOTA performance among perception annotation-free approaches, demonstrating a 21\% reduction in L2 error (0.61 m → 0.48 m) and an impressive 83\% reduction in collision rate (0.30\% → 0.05\%) compared to the strong baseline LAW. Moreover, WorldRFT achieves the lowest collision rate among all evaluated methods, even surpassing perception-based approaches, which underscores the superior safety performance of our planning-oriented design.
We also provide results incorporating the ego status, with specific descriptions detailed in the supplementary materials.

In addition, as demonstrated in Table~\ref{tab2}, WorldRFT also achieves SOTA performance in closed-loop evaluation with PDMS of 87.8 using only camera inputs. Compared to the baseline LAW (84.6), our approach demonstrates significant improvements across all safety-critical metrics: NC (No At-fault Collision) (96.4 → 97.8), TTC (Time-to-Collision) (88.7 → 94.0), and particularly DAC (Drivable Area Compliance) (95.4 → 96.8). The exceptional DAC score of 96.8, which is the highest among all methods including LiDAR-based approaches, demonstrates that incorporating VGGT geometric priors substantially enhances the model's 3D spatial understanding without explicit depth supervision.

Notably, our closed-loop performance exceeds that of most methods relying on LiDAR inputs, trailing DiffusionDrive (88.1) by only 0.3 points. This result demonstrates that our planning-oriented latent world modeling, combined with safety-oriented reinforcement fine-tuning, achieves outstanding performance and holds strong potential for real-world deployment.

\subsection{Ablation Study}

In this section, we conduct comprehensive ablation studies to investigate the effectiveness of each component in our proposed WorldRFT framework. All ablation experiments incorporate our safety-oriented reinforcement fine-tuning by default to ensure consistent safety-aware evaluation. All of the ablation experiments are conducted on the nuScenes and NavSim benchmark. 
For clarity, we present detailed results on nuScenes here, while NavSim results are provided in the supplementary material.

\textbf{Effectiveness of VGGT Spatial Encoding.}
As demonstrated in Table~\ref{tab:ablation}, incorporating VGGT geometric priors shows consistent improvements across different configurations. Comparing ID 4 with ID 8, adding VGGT reduces L2 error by 7.7\% (0.52 → 0.48) and collision rate by 37.5\% (0.08 → 0.05). The performance gains remain substantial even with partial task designs (ID 5-6), demonstrating that VGGT-enhanced spatial understanding benefits planning regardless of the specific task configuration.

\textbf{Impact of Hierarchical Planning Tasks.} 
Starting from the baseline (ID 1 in Table~\ref{tab:ablation}), introducing target region localization and spatial path planning with refinement (ID 4) reduces L2 error by 11.9\% (0.59 → 0.52) and collision rate by 50.0\% (0.16 → 0.08). This validates our hypothesis that decomposing the complex planning task into hierarchical subtasks enables more effective feature extraction from latent representations, thereby enhancing overall planning performance.

\textbf{Local-aware Iterative Refinement.} 
As shown in Table~\ref{tab:ablation}, comparing ID 7 (without refinement) and ID 8 (with refinement), the refinement module reduces L2 error by 4.0\% (0.50 → 0.48) and collision rate by 16.7\% (0.06 → 0.05). This demonstrates that incorporating local features based on initial trajectories enables fine-grained trajectory adjustments that improve both trajectory accuracy and safety.

\textbf{Safety-Oriented Reinforcement Fine-tuning.} As shown in Table~\ref{tab1}, comparing our method without RFT (Ours without RFT) and with RFT (Ours with RFT), the RL module significantly reduces collision rate by 66.7\% (0.15 → 0.05) while L2 error slightly increases from 0.47 to 0.48. This selective improvement in safety metrics validates that our collision-aware reward design guides the model from behavior imitation toward understanding safety principles.

\begin{table}[t]
    \centering
    \small
    {
    \begin{tabular}{c|c|cc|c|c|c}
        \toprule
        \multirow{2}{*}{ID} & \multicolumn{1}{c|}{Latent Encoder} &  \multicolumn{2}{c|}{Planning Task} & \multirow{2}{*}{\makecell{Refine\\-ment}} & \multirow{2}{*}{L2} & \multirow{2}{*}{CR} \\
        \cline{2-4}
        & VGGT & Target & Path & & & \\ 
        \midrule
        1 & & & & & 0.59 & 0.16 \\  
        2 & & \checkmark & &\checkmark & 0.56 & 0.10 \\
        3 & & & \checkmark &\checkmark & 0.54 & 0.11 \\
        4 & & \checkmark & \checkmark &\checkmark & 0.52 & 0.08 \\
        5 & \checkmark & & \checkmark &\checkmark & 0.51 & 0.08 \\
        6 & \checkmark & \checkmark & &\checkmark & 0.49 & 0.09 \\
        7 & \checkmark & \checkmark & \checkmark & & 0.50 & 0.06 \\
        8 & \checkmark & \checkmark & \checkmark &\checkmark & 0.48 & 0.05 \\
        \bottomrule
    \end{tabular}
    }
    \caption{Ablation study of each proposed component}
    \label{tab:ablation}
\end{table}

\begin{figure}[!t]
    \centering
    \includegraphics[width=1\columnwidth]{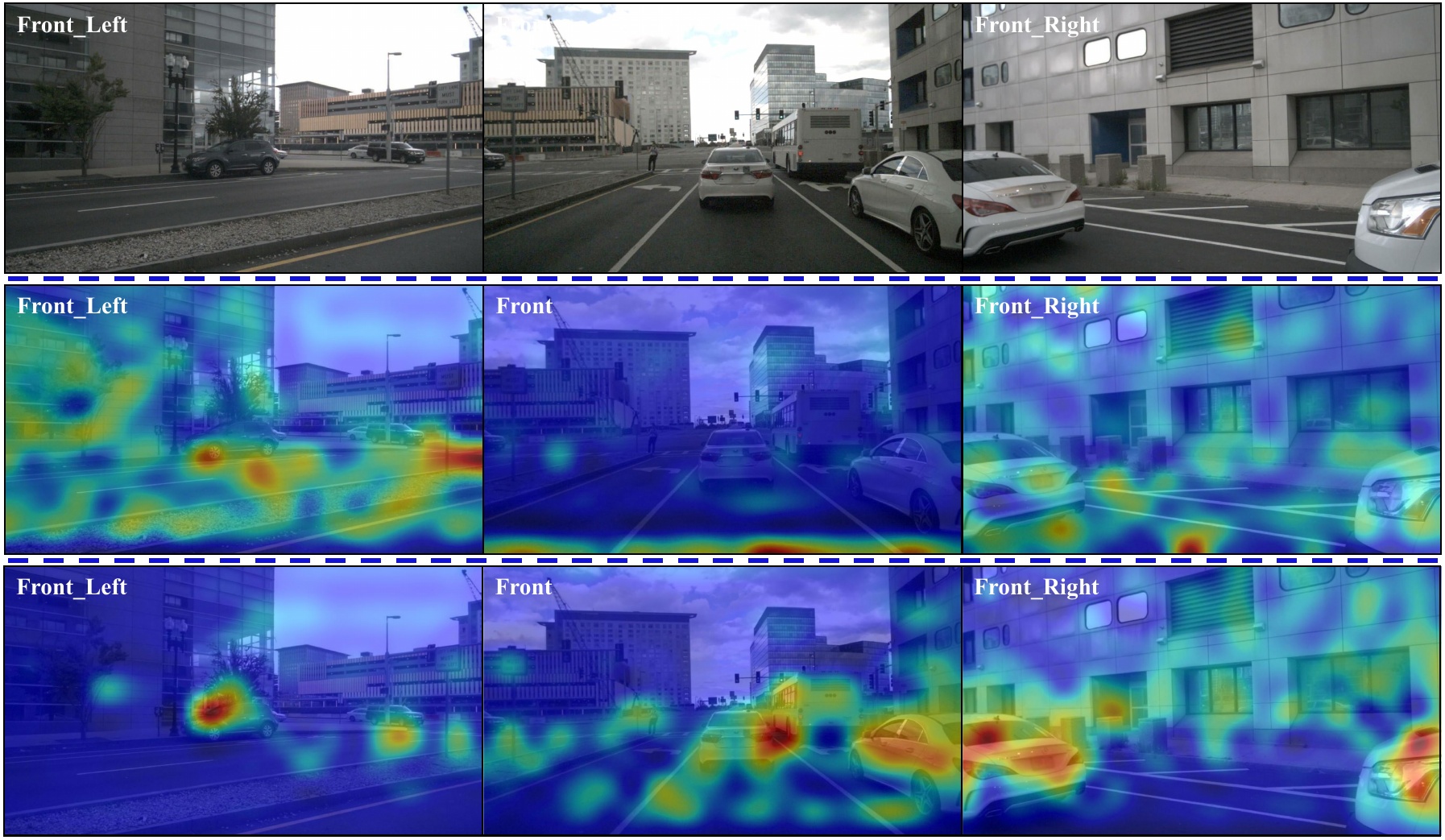}
    \caption{Visualization of the original image (top), along with attention maps for the baseline (middle) and WorldRFT (bottom), where blue represents weak attention and red indicates strong attention. The maps reveal that WorldRFT focuses on surrounding planning-critical agents such as vehicles, which are largely overlooked by the baseline method.}
    \label{fig:4}
\end{figure}

\textbf{Qualitative Results}
In this section, we quantitatively evaluate WorldRFT on nuScenes. Figure~\ref{fig:3} shows predicted trajectories: ground truth (red), without RFT (blue), and with RFT (green). The without-RFT model approaches surrounding vehicles too closely, while the with-RFT model maintains safer distances and better matches the ground truth. As shown in Figure~\ref{fig:4}, attention maps (LAW middle, WorldRFT below) demonstrate that WorldRFT focuses more precisely on critical regions like surrounding vehicles, improving perception-planning integration and yielding more robust, human-like behavior.

\section{Conclusion}

We present WorldRFT, a planning-oriented latent world framework for end-to-end autonomous driving. Our key insight is that existing latent world models suffer from misalignment between reconstruction-oriented learning and planning requirements. To bridge this gap, we introduced three innovations: 
(1) SWE with VGGT for spatial-aware representations; 
(2) HPR combining hierarchical task decomposition with iterative refinement for planning-relevant features; and 
(3) RFT enabling  active collision avoidance beyond passive imitation. 
Experiments on nuScenes and NavSim demonstrate that our planning-oriented design with geometric priors and safety optimization effectively enables slight end-to-end autonomous driving. 



\bibliography{aaai2026.bib}

\begin{thebibliography}{36}
\providecommand{\natexlab}[1]{#1}

\bibitem[{Caesar et~al.(2020)Caesar, Bankiti, Lang, Vora, Liong, Xu, Krishnan, Pan, Baldan, and Beijbom}]{caesar2020nuscenes}
Caesar, H.; Bankiti, V.; Lang, A.~H.; Vora, S.; Liong, V.~E.; Xu, Q.; Krishnan, A.; Pan, Y.; Baldan, G.; and Beijbom, O. 2020.
\newblock nuScenes: A Multimodal dataset for autonomous driving.
\newblock In \emph{Proceedings of the IEEE/CVF Conference on Computer Vision and Pattern Recognition}, 11621--11631.

\bibitem[{Chen et~al.(2024{\natexlab{a}})Chen, Jiang, Gao, Liao, Xu, Zhang, Huang, Liu, and Wang}]{chen2024vadv2}
Chen, S.; Jiang, B.; Gao, H.; Liao, B.; Xu, Q.; Zhang, Q.; Huang, C.; Liu, W.; and Wang, X. 2024{\natexlab{a}}.
\newblock Vadv2: End-to-end vectorized autonomous driving via probabilistic planning.
\newblock \emph{arXiv preprint arXiv:2402.13243}.

\bibitem[{Chen et~al.(2024{\natexlab{b}})Chen, Ye, Xu, Cao, and Chen}]{chen2024ppad}
Chen, Z.; Ye, M.; Xu, S.; Cao, T.; and Chen, Q. 2024{\natexlab{b}}.
\newblock Ppad: Iterative interactions of prediction and planning for end-to-end autonomous driving.
\newblock In \emph{Proceedings of the European Conference on Computer Vision}, 239--256. Springer.

\bibitem[{Dauner et~al.(2024)Dauner, Hallgarten, Li, Weng, Huang, Yang, Li, Gilitschenski, Ivanovic, Pavone, Geiger, and Chitta}]{im2024navsim}
Dauner, D.; Hallgarten, M.; Li, T.; Weng, X.; Huang, Z.; Yang, Z.; Li, H.; Gilitschenski, I.; Ivanovic, B.; Pavone, M.; Geiger, A.; and Chitta, K. 2024.
\newblock NAVSIM: Data-driven non-reactive autonomous vehicle simulation and benchmarking.
\newblock \emph{arXiv preprint arXiv:2406.15349}.

\bibitem[{Gao et~al.(2025)Gao, Chen, Jiang, Liao, Shi, Guo, Pu, Yin, Li, Zhang, Zhang, Liu, Zhang, and Wang}]{gao2025rad}
Gao, H.; Chen, S.; Jiang, B.; Liao, B.; Shi, Y.; Guo, X.; Pu, Y.; Yin, H.; Li, X.; Zhang, X.; Zhang, Y.; Liu, W.; Zhang, Q.; and Wang, X. 2025.
\newblock Rad: Training an end-to-end driving policy via large-scale 3dgs-based reinforcement learning.
\newblock \emph{arXiv preprint arXiv:2502.13144}.

\bibitem[{Gao et~al.(2024)Gao, Zhang, Ding, and Zhao}]{10547289}
Gao, Y.; Zhang, Q.; Ding, D.-W.; and Zhao, D. 2024.
\newblock Dream to Drive With Predictive Individual World Model.
\newblock \emph{IEEE Transactions on Intelligent Vehicles}, 9(12): 8224--8238.

\bibitem[{Gu et~al.(2024)Gu, Yin, Jin, Guo, Wang, Li, Zhang, and Long}]{gu2024dome}
Gu, S.; Yin, W.; Jin, B.; Guo, X.; Wang, J.; Li, H.; Zhang, Q.; and Long, X. 2024.
\newblock Dome: Taming diffusion model into high-fidelity controllable occupancy world model.
\newblock \emph{arXiv preprint arXiv:2410.10429}.

\bibitem[{Hu et~al.(2023{\natexlab{a}})Hu, Russell, Yeo, Murez, Fedoseev, Kendall, Shotton, and Corrado}]{hu2023gaia}
Hu, A.; Russell, L.; Yeo, H.; Murez, Z.; Fedoseev, G.; Kendall, A.; Shotton, J.; and Corrado, G. 2023{\natexlab{a}}.
\newblock Gaia-1: A generative world model for autonomous driving.
\newblock \emph{arXiv preprint arXiv:2309.17080}.

\bibitem[{Hu et~al.(2024)Hu, Yin, Zhang, Cai, Long, Chen, Wang, Yu, Shen, and Shen}]{hu2024metric3d}
Hu, M.; Yin, W.; Zhang, C.; Cai, Z.; Long, X.; Chen, H.; Wang, K.; Yu, G.; Shen, C.; and Shen, S. 2024.
\newblock Metric3d v2: A versatile monocular geometric foundation model for zero-shot metric depth and surface normal estimation.
\newblock \emph{IEEE Transactions on Pattern Analysis and Machine Intelligence}.

\bibitem[{Hu et~al.(2022)Hu, Chen, Wu, Li, Yan, and Tao}]{hu2022st}
Hu, S.; Chen, L.; Wu, P.; Li, H.; Yan, J.; and Tao, D. 2022.
\newblock St-p3: End-to-end vision-based autonomous driving via spatial-temporal feature learning.
\newblock In \emph{Proceedings of the European Conference on Computer Vision}, 533--549. Springer.

\bibitem[{Hu et~al.(2023{\natexlab{b}})Hu, Yang, Chen, Li, Sima, Zhu, Chai, Du, Lin, Wang, Lu, Jia, Liu, Dai, Qiao, and Li}]{hu2023planning}
Hu, Y.; Yang, J.; Chen, L.; Li, K.; Sima, C.; Zhu, X.; Chai, S.; Du, S.; Lin, T.; Wang, W.; Lu, L.; Jia, X.; Liu, Q.; Dai, J.; Qiao, Y.; and Li, H. 2023{\natexlab{b}}.
\newblock Planning-oriented autonomous driving.
\newblock In \emph{Proceedings of the IEEE/CVF Conference on Computer Vision and Pattern Recognition}, 17853--17862.

\bibitem[{Jia et~al.(2025)Jia, You, Zhang, and Yan}]{jia2025drivetransformer}
Jia, X.; You, J.; Zhang, Z.; and Yan, J. 2025.
\newblock Drivetransformer: Unified transformer for scalable end-to-end autonomous driving.
\newblock In \emph{Proceedings of the International Conference on Learning Representations}.

\bibitem[{Jiang et~al.(2023)Jiang, Chen, Xu, Liao, Chen, Zhou, Zhang, Liu, Huang, and Wang}]{jiang2023vad}
Jiang, B.; Chen, S.; Xu, Q.; Liao, B.; Chen, J.; Zhou, H.; Zhang, Q.; Liu, W.; Huang, C.; and Wang, X. 2023.
\newblock VAD: Vectorized scene representation for efficient autonomous driving.
\newblock \emph{Proceedings of the IEEE/CVF International Conference on Computer Vision}.

\bibitem[{Jiang et~al.(2025)Jiang, Chen, Zhang, Liu, and Wang}]{jiang2025alphadrive}
Jiang, B.; Chen, S.; Zhang, Q.; Liu, W.; and Wang, X. 2025.
\newblock AlphaDrive: Unleashing the power of VLMs in autonomous driving via reinforcement learning and reasoning.
\newblock \emph{arXiv preprint arXiv:2503.07608}.

\bibitem[{Jin et~al.(2025)Jin, Gu, Hu, Zheng, Guo, Zhang, Long, and Yin}]{jin2025occtens}
Jin, B.; Gu, S.; Hu, X.; Zheng, Y.; Guo, X.; Zhang, Q.; Long, X.; and Yin, W. 2025.
\newblock OccTENS: 3D Occupancy World Model via Temporal Next-Scale Prediction.
\newblock \emph{arXiv preprint arXiv:2509.03887}.

\bibitem[{Li and Cui(2025)}]{li2025navigationguided}
Li, P.; and Cui, D. 2025.
\newblock Navigation-guided sparse scene representation for end-to-end autonomous driving.
\newblock In \emph{Proceedings of the International Conference on Learning Representations}, 15522--15533.

\bibitem[{Li et~al.(2025)Li, Li, Zheng, Sun, Wang, and Chen}]{li2025semi}
Li, X.; Li, P.; Zheng, Y.; Sun, W.; Wang, Y.; and Chen, Y. 2025.
\newblock Semi-supervised vision-centric 3d occupancy world model for autonomous driving.
\newblock \emph{arXiv preprint arXiv:2502.07309}.

\bibitem[{Li, Fan et~al.(2025)}]{li2024enhancing}
Li, Y.; Fan, L.; et~al. 2025.
\newblock Enhancing end-to-end autonomous driving with latent world model.
\newblock \emph{Proceedings of the International Conference on Learning Representations}.

\bibitem[{Li et~al.(2024)Li, Li, Wang, Lan, Yu, Ji, Li, Zhu, Kautz, Wu, Jiang, and Alvarez}]{li2024hydra}
Li, Z.; Li, K.; Wang, S.; Lan, S.; Yu, Z.; Ji, Y.; Li, Z.; Zhu, Z.; Kautz, J.; Wu, Z.; Jiang, Y.; and Alvarez, J.~M. 2024.
\newblock Hydra-mdp: End-to-end multimodal planning with multi-target hydra-distillation.
\newblock \emph{arXiv preprint arXiv:2406.06978}.

\bibitem[{Liao, Chen et~al.(2025)}]{liao2024diffusiondrive}
Liao, B.; Chen, S.; et~al. 2025.
\newblock Diffusiondrive: Truncated diffusion model for end-to-end autonomous driving.
\newblock \emph{Proceedings of the IEEE/CVF Conference on Computer Vision and Pattern Recognition}.

\bibitem[{Min et~al.(2024)Min, Zhao, Xiao, Zhao, Xu, Zhu, Jin, Li, Guo, Xing, Jing, Nie, and Dai}]{min2024driveworld}
Min, C.; Zhao, D.; Xiao, L.; Zhao, J.; Xu, X.; Zhu, Z.; Jin, L.; Li, J.; Guo, Y.; Xing, J.; Jing, L.; Nie, Y.; and Dai, B. 2024.
\newblock Driveworld: 4D pre-trained scene understanding via world models for autonomous driving.
\newblock In \emph{Proceedings of the IEEE/CVF Conference on Computer Vision and Pattern Recognition}, 15522--15533.

\bibitem[{Prakash, Chitta, and Geiger(2021)}]{prakash2021multi}
Prakash, A.; Chitta, K.; and Geiger, A. 2021.
\newblock Multi-modal fusion transformer for end-to-end autonomous driving.
\newblock In \emph{Proceedings of the IEEE/CVF Conference on Computer Vision and Pattern Recognition}.

\bibitem[{Ren et~al.(2024)Ren, Liu, Zeng, Lin, Li, Cao, Chen, Huang, Chen, Yan, Zeng, Zhang, Li, Yang, Li, Jiang, and Zhang}]{ren2024grounded}
Ren, T.; Liu, S.; Zeng, A.; Lin, J.; Li, K.; Cao, H.; Chen, J.; Huang, X.; Chen, Y.; Yan, F.; Zeng, Z.; Zhang, H.; Li, F.; Yang, J.; Li, H.; Jiang, Q.; and Zhang, L. 2024.
\newblock Grounded sam: Assembling open-world models for diverse visual tasks.
\newblock \emph{arXiv preprint arXiv:2401.14159}.

\bibitem[{Shao et~al.(2024)Shao, Wang, Zhu, Xu, Song, Bi, Zhang, Zhang, Li, and Guo}]{shao2024grpo}
Shao, Z.; Wang, P.; Zhu, Q.; Xu, R.; Song, J.; Bi, X.; Zhang, H.; Zhang, M.; Li, Y., Y.K.and~Wu; and Guo, D. 2024.
\newblock DeepSeekMath: Pushing the limits of mathematical reasoning in open language models.
\newblock \emph{arXiv preprint arXiv:2402.03300}.

\bibitem[{Shi et~al.(2025)Shi, Shi, Sheng, Zhang, and Jiang}]{shi2025drivex}
Shi, C.; Shi, S.; Sheng, K.; Zhang, B.; and Jiang, L. 2025.
\newblock DriveX: Omni scene modeling for learning generalizable world knowledge in autonomous driving.
\newblock \emph{arXiv preprint arXiv:2505.19239}.

\bibitem[{Sun et~al.(2024)Sun, Lin, Shi, Zhang, Wu, and Zheng}]{sun2024sparsedrive}
Sun, W.; Lin, X.; Shi, Y.; Zhang, C.; Wu, H.; and Zheng, S. 2024.
\newblock Sparsedrive: End-to-end autonomous driving via sparse scene representation.
\newblock \emph{arXiv preprint arXiv:2405.19620}.

\bibitem[{Tong et~al.(2023)Tong, Sima, Wang, Chen, Wu, Deng, Gu, Lu, Luo, Lin, and H.}]{tong2023scene}
Tong, W.; Sima, C.; Wang, T.; Chen, L.; Wu, S.; Deng, H.; Gu, Y.; Lu, L.; Luo, P.; Lin, D.; and H., L. 2023.
\newblock Scene as occupancy.
\newblock In \emph{Proceedings of the IEEE/CVF International Conference on Computer Vision}, 8406--8415.

\bibitem[{Wang et~al.(2025)Wang, Chen, Karaev, Vedaldi, Rupprecht, and Novotny}]{wang2025vggt}
Wang, J.; Chen, M.; Karaev, N.; Vedaldi, A.; Rupprecht, C.; and Novotny, D. 2025.
\newblock Vggt: Visual geometry grounded transformer.
\newblock In \emph{Proceedings of the Computer Vision and Pattern Recognition Conference}, 5294--5306.

\bibitem[{Weng et~al.(2024)Weng, Ivanovic, Wang, Wang, and Pavone}]{weng2024para}
Weng, X.; Ivanovic, B.; Wang, Y.; Wang, Y.; and Pavone, M. 2024.
\newblock PARA-Drive: Parallelized architecture for real-time autonomous driving.
\newblock In \emph{Proceedings of the IEEE/CVF Conference on Computer Vision and Pattern Recognition}, 15449--15458.

\bibitem[{Xing et~al.(2025)Xing, Zhang, Hu, Jiang, He, Zhang, Long, and Yin}]{xing2025goalflow}
Xing, Z.; Zhang, X.; Hu, Y.; Jiang, B.; He, T.; Zhang, Q.; Long, X.; and Yin, W. 2025.
\newblock Goalflow: Goal-driven flow matching for multimodal trajectories generation in end-to-end autonomous driving.
\newblock In \emph{Proceedings of the Computer Vision and Pattern Recognition Conference}, 1602--1611.

\bibitem[{Yang et~al.(2025)Yang, Zheng, Zhang, Zhu, Xing, Lin, Liu, Su, and Zhao}]{yang2025uncad}
Yang, P.; Zheng, Y.; Zhang, Q.; Zhu, K.; Xing, Z.; Lin, Q.; Liu, Y.-F.; Su, Z.; and Zhao, D. 2025.
\newblock UncAD: Towards Safe End-to-end Autonomous Driving via Online Map Uncertainty.
\newblock \emph{arXiv preprint arXiv:2504.12826}.

\bibitem[{Zhang et~al.(2025)Zhang, Tang, Hu, Pan, Guo, Liu, Huang, Yuan, Zhang, Long, Cao, and Yin}]{zhang2025epona}
Zhang, K.; Tang, Z.; Hu, X.; Pan, X.; Guo, X.; Liu, Y.; Huang, J.; Yuan, L.; Zhang, Q.; Long, X.; Cao, X.; and Yin, W. 2025.
\newblock Epona: Autoregressive diffusion world model for autonomous driving.
\newblock \emph{arXiv preprint arXiv:2506.24113}.

\bibitem[{Zhao et~al.(2025)Zhao, Ni, Wang, Zhu, Zhang, Wang, Huang, Chen, Wang, Zhang, Mei, and Wang}]{zhao2025drivedreamer4D}
Zhao, G.; Ni, C.; Wang, X.; Zhu, Z.; Zhang, X.; Wang, Y.; Huang, G.; Chen, X.; Wang, B.; Zhang, Y.; Mei, W.; and Wang, X. 2025.
\newblock DriveDreamer4D: World models are effective data machines for 4d driving scene representation.
\newblock In \emph{Proceedings of the IEEE/CVF Conference on Computer Vision and Pattern Recognition}, 12015--12026.

\bibitem[{Zheng et~al.(2024{\natexlab{a}})Zheng, Song, Guo, Zhang, and Chen}]{zheng2024genad}
Zheng, W.; Song, R.; Guo, X.; Zhang, C.; and Chen, L. 2024{\natexlab{a}}.
\newblock Genad: Generative end-to-end autonomous driving.
\newblock In \emph{European Conference on Computer Vision}, 87--104. Springer.

\bibitem[{Zheng et~al.(2024{\natexlab{b}})Zheng, Xia, Zhang, Zhang, Lu, Huo, Han, Li, Yu, Jin et~al.}]{zheng2024preliminary}
Zheng, Y.; Xia, Z.; Zhang, Q.; Zhang, T.; Lu, B.; Huo, X.; Han, C.; Li, Y.; Yu, M.; Jin, B.; et~al. 2024{\natexlab{b}}.
\newblock Preliminary investigation into data scaling laws for imitation learning-based end-to-end autonomous driving.
\newblock \emph{arXiv preprint arXiv:2412.02689}.

\bibitem[{Zheng et~al.(2025)Zheng, Yang, Xing, Zhang, Zheng, Gao, Li, Zhang, Xia, Jia, and Zhao}]{zheng2025world4drive}
Zheng, Y.; Yang, P.; Xing, Z.; Zhang, Q.; Zheng, Y.; Gao, Y.; Li, P.; Zhang, T.; Xia, Z.; Jia, P.; and Zhao, D. 2025.
\newblock World4drive: End-to-end autonomous driving via intention-aware physical latent world model.
\newblock \emph{arXiv preprint arXiv:2507.00603}.

\end{thebibliography}

\newpage
\twocolumn[ 
  \begin{center}
    \Huge\bfseries WorldRFT: Latent World Model Planning with Reinforcement Fine-Tuning for Autonomous Driving
    
    \vspace{1em}
    
    \Large Supplementary Material
  \end{center}
  \vspace{2em}
]



\section{Implementation Details}
In this section, we provide comprehensive implementation details for WorldRFT, including the hierarchical planning strategy and training configurations for both nuScenes and NavSim benchmarks. These details complement the main paper and enable reproducibility of our results.

\subsection{Hierarchical Planning }
Our hierarchical planning framework decomposes the complex trajectory generation task into three parallel subtasks that capture distinct aspects of planning: 
\begin{itemize}
    \item Target region localization models future destinations as probabilistic Laplace distributions rather than deterministic points, capturing the inherent uncertainty in driving intentions. Thus, the ground-truth for the target region center is provided by the coordinates of the final position of the ground-truth trajectory. 
    \item Spatial path planning produces a geometric path represented by a set of time-independent, equally spaced points that capture only the spatial coordinates of the trajectory. Ground-truth supervision is generated through uniform interpolation of the ground-truth trajectory. This approach streamlines learning by allowing the model to focus exclusively on spatial direction, thereby enhancing its understanding of spatial coordinates.
    \item Temporal trajectory prediction further refines the geometric path by incorporating temporal dynamics at fixed time intervals. Here, the trajectory is represented as a sequence of time-stamped points, capturing both spatial and dynamic (velocity and acceleration) information. Ground-truth supervision is obtained by uniformly sampling the ground-truth trajectory over time, allowing the model to learn spatial and temporal aspects of motion.
\end{itemize}

These subtasks are unified through a query interaction framework where dedicated queries ($Q_{\text{target}}$, $Q_{\text{path}}$, $Q_{\text{traj}}$) first extract task-specific features via cross-attention with the latent world representation, then exchange information through self-attention to achieve mutual awareness of planning intentions. This hierarchical decomposition not only simplifies the learning process but also enables more interpretable and controllable trajectory generation.

\subsection{NuScenes Training}
Following the VAD-Tiny~\cite{jiang2023vad} configuration, we set the default feature dimension to 256 and employ ResNet-50 as the image backbone to process 6 surround-view images with a resolution of $360 \times 640$. When utilizing VGGT, which requires a maximum image input width of 518, we first rescale the original images and then crop them to $294 \times 518$ before feeding them into the VGGT module.

Following World4Drive~\cite{zheng2025world4drive}, we incorporate Grounded-SAM~\cite{ren2024grounded} for semantic learning. Specifically, we utilize this vision-language model to produce pseudo semantic labels. Given the prompt for the object of interest, we obtain 2D bounding boxes and the corresponding semantic mask $S_t \in \mathbb{R}^{H \times W \times C}$ via the Grounded-SAM model, where we only keep labels with high confidence to reduce incorrect labeling. Finally, we leverage cross-entropy loss $\mathcal{L}_{sem}$ to enhance the semantic understanding of the latent representations.

The driving commands are consistent with previous work~\cite{hu2023planning}, comprising three categories: left, right, and straight. For the spatial path, we set a fixed distance of 2 m between consecutive trajectory points with a total of 30 points. For the temporal trajectory, following previous work~\cite{sun2024sparsedrive}, we predict future trajectories for 3 seconds with a time interval of 0.5 seconds, yielding 6 trajectory points in total. 
As for the target region center, we use the final position of the ground-truth trajectory as the ground-truth label for supervision.

For the trajectory refinement module, we set the number of iterations $K$ to 3. The model is trained for 12 epochs on 8 NVIDIA RTX 3090 GPUs with a total batch size of 8 and an initial learning rate of $5 \times 10^{-5}$. The loss weights are set as 0.2, 0.2, 0.001, and 1.0 for semantic, reconstruction, target, and trajectory losses, respectively. 

For reinforcement fine-tuning, we utilize the pre-trained model output as the mean and apply the GRPO strategy for optimization with a variance network. To better align the model's behavior with that of the reference model, we introduce an additional reference loss into the reinforcement learning framework, which can be formulated as:
\begin{equation}
   \mathcal{L}_\text{ref}=\frac{1}{BT}\sum^{B}\sum^{T}\Vert \boldsymbol{\mu}_{\theta}-\boldsymbol{\mu}_{ref}\Vert_2
\end{equation}
where $B$ and $T$ denote the batch size and the time step, respectively. $\boldsymbol{\mu}_{\theta}$ and $\boldsymbol{\mu}_{ref}$ represent the mean trajectories output by the current model and the reference model, respectively. Last but not least, we incorporate a maximum entropy loss for the normal distribution to prevent premature convergence during reinforcement learning, thereby ensuring the effectiveness of sampling. The initial learning rate is set to $3 \times 10^{-6}$, with a group number of 10. The coefficients for the KL loss, reference loss, and maximum entropy loss are set to 0.1, 0.12, and 0.1, respectively.

\begin{figure*}[!htbp]
    \centering
    \includegraphics[width =1\linewidth]{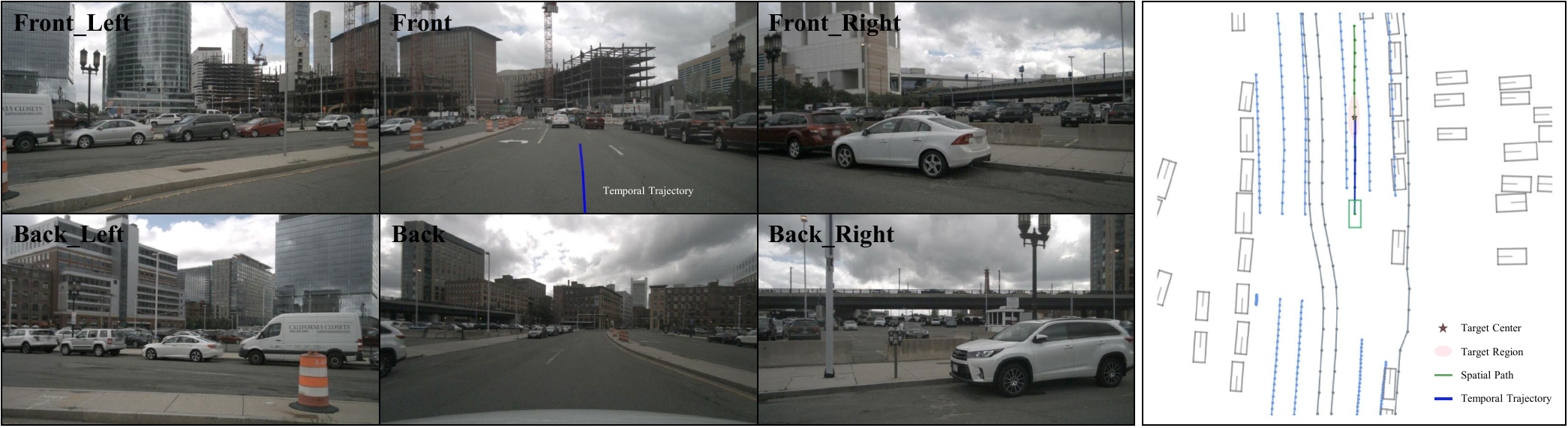}
    \caption{Visualization of WorldRFT on nuScenes in a straight driving scenario. Left: Input images; Right: Predicted target region (ellipse), spatial path, and trajectory. The narrow longitudinal ellipse indicates high uncertainty in speed but confident directional prediction.}
    \label{fig:1}
\end{figure*}

\begin{figure*} [!htbp]
    \centering
    \includegraphics[width =1\linewidth]{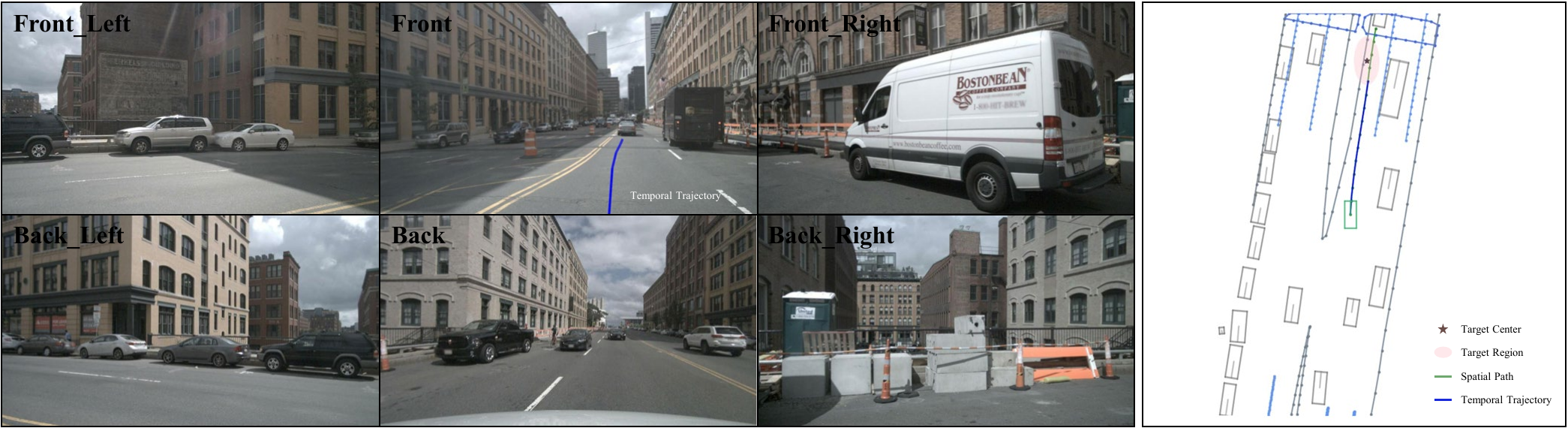}
    \caption{Visualization of WorldRFT on nuScenes in a scenario with an inaccurate target region center. Left: Input images; Right: Predicted target region (ellipse), spatial path, and trajectory. The large ellipse indicates high uncertainty in both speed and direction. Despite center inaccuracy, the probabilistic target region allows for cautious, safety-aware decisions.}
    \label{fig:2}
\end{figure*}

\subsection{NavSim Training}
In alignment with the NavSim benchmark, our closed-loop model takes as input a concatenated image formed by stitching the front, front-left, and front-right camera views, which is subsequently resized to $256 \times 1024$. We employ ResNet-34 for image feature extraction to balance computational efficiency with model performance.

For VGGT and Grounded-SAM, we follow the same usage methodology as described in our NuScenes training approach. Specifically, we apply the same preprocessing pipeline for VGGT input and utilize Grounded-SAM for semantic pseudo-label generation with identical confidence thresholds.

Regarding trajectory configuration, for the spatial path, we set 50 points with a fixed interval of 2 m between consecutive points. For the temporal trajectory, consistent with other methods~\cite{xing2025goalflow} in NavSim, we predict future trajectories for 4 seconds with a time interval of 0.5 seconds, resulting in 8 trajectory points in total. 

The model is trained for 100 epochs on 8 NVIDIA RTX 3090 GPUs with a batch size of 64. For reinforcement learning, we adopt the same reward design and experimental setup as in our NuScenes training. The reinforcement learning phase is conducted for 10 epochs on 8 NVIDIA RTX 3090 GPUs, maintaining the same batch size of 64.





\section{Dataset Introduction}
\subsection{nuScenes Benchmark}
\subsubsection{Overview.}
The nuScenes dataset includes 1,000 selected 20-second driving sequences—approximately 15 hours of real-world data—collected in Boston and Singapore \cite{caesar2020nuscenes}. It captures diverse driving scenarios, complex traffic conditions, and rare events, offering high realism. The dataset provides rich multimodal sensor data: about 1.4 million camera images, 390,000 LiDAR sweeps, 1.4 million radar sweeps, and 1.4 million annotated 3D bounding boxes across 40,000 keyframes, making it a comprehensive resource for training and evaluating autonomous driving systems.

\subsubsection{Detailed Metrics.}
The nuScenes dataset employs L2 Error and Collision Rate as key metrics for trajectory evaluation. The L2 Error quantifies the average Euclidean distance between the predicted (planned) trajectory and the actual trajectory driven by a human. The Collision Rate reflects the frequency with which the planned trajectory results in collisions with other road agents, serving as a critical indicator of safety and planning reliability.

\subsubsection{Data Processing Pipeline.}
For the nuScenes experiments, the processing of the six surrounding camera images begins with resizing the original 900 × 1600 frames to 360 × 640. These are then uniformly padded to achieve a standardized input dimension of 384 × 640, forming the batched input tensor ${I_t} \in {\mathbb{R}^{6 \times 3 \times 384 \times 640}}$. The image features are extracted using a backbone network, which applies spatial downsampling to reduce the resolution to 12 × 20. The resulting high-level feature map, represented as  ${F_t} \in {\mathbb{R}^{6 \times 240 \times 256}}$, captures rich contextual information and is used as input for downstream network components.

\subsection{NavSim Benchmark}
\subsubsection{Overview.}
The NavSim benchmark leverages the OpenScenedataset, comprising 1,192 training sequences and 136 test sequences, collectively containing more than 100,000 keyframes \cite{im2024navsim}. Following the official baseline protocol, the model outputs trajectories at a frequency of 2 Hz over a 4-second planning horizon, which are then refined and upsampled using a Linear Quadratic Regulator (LQR) controller to generate smooth, dynamically feasible motion plans.
\subsubsection{Detailed Metrics.}
NavSim employs the PDM Score (PDMS) as the primary metric for trajectory evaluation, with the specific scoring formula defined as:
\begin{equation}
    \text{PDMS}=\text{NC}\times\text{DAC} \\
    \times \frac{(5\times\text{EP}+5\times\text{TTC}+2\times\text{Comf.})}{12}
\end{equation}
where NC denotes no at-fault collision, DAC represents drivable area compliance, TTC is time-to-collision, Comf. indicates ride comfort, and EP stands for ego progress. NC and DAC act as multiplicative factors in the evaluation, penalizing unsafe or infeasible trajectories. NC measures the likelihood of collisions between the ego vehicle and other agents, while DAC assesses whether the predicted path remains within drivable areas. Other metrics are combined via weighted summation. EP quantifies the expected progress along the route over the next 4 seconds. TTC evaluates immediate safety by estimating the time to potential collision under constant motion. Comf. assesses driving smoothness by analyzing acceleration, heading changes, and other dynamic behaviors.

\subsubsection{Data Processing Pipeline.}
In the NavSim experiments, the input comprises ego state $e_t = [c_t,v_t,a_t]$ and image data $I_t \in \mathbb{R}^{3 \times 256 \times 1024}$, where $c_t$ is a one-hot vector encoding the navigation command (left, straight, right), and $v_t$ and $a_t$ represent the ego vehicle's velocity and acceleration at time $t$, respectively. The raw image input  $I^{\text{raw}}_t \in \mathbb{R}^{3 \times 1080 \times 1920}$ is preprocessed as follows: the front view $I^{\text{front}}_t$ is cropped vertically to $\mathbb{R}^{3 \times 1024 \times 1920}$, while left $I^{\text{left}}_t$ and right views $I^{\text{right}}_t$ are cropped in both height and width to $\mathbb{R}^{3 \times 1024 \times 1088}$. These are then concatenated along the width axis to form:
\begin{equation}
    I^{\text{concat}}_t = \text{Concat}(I^{\text{left}}_t, I^{\text{front}}_t, I^{\text{right}}_t)
\end{equation}

Finally, the concatenated image is resized to the final input resolution: $\mathbb{R}^{3 \times 256 \times 1024}$.



\section{Additional Experimental Results}
This section presents comprehensive ablation studies that systematically evaluate each component's contribution and analyze hyperparameters' impact on model performance.

\subsection{Ablation study on NavSim}
Table~\ref{tab:ablation} presents the ablation study of each proposed component on the NavSim benchmark. We systematically evaluate the contribution of VGGT latent encoder, hierarchical planning tasks (target prediction and path planning), and the refinement module. 
All ablation experiments incorporate our safety-oriented reinforcement fine-tuning by default to ensure consistent safety-aware evaluation.
The results demonstrate that each component contributes to the overall performance improvement, with the complete model achieving the highest PDMS of 87.8. The most significant improvement comes from incorporating VGGT (ID 4 → ID 5: +0.7 PDMS), validating the importance of spatial-aware representations. The hierarchical planning tasks (target and path) provide complementary benefits, with their combination yielding better results than individual tasks. The refinement module adds the final performance boost (ID 7 → ID 8: +0.2 PDMS), demonstrating the value of local-aware iterative refinement.

\begin{table}[t]
    \centering
    \small
    \caption{Ablation study of each proposed component on NavSim benchmark. Components are systematically removed to evaluate individual contributions.}
    {
    \begin{tabular}{c|c|cc|c|c}
        \toprule
        \multirow{2}{*}{ID} & \multicolumn{1}{c|}{Latent Encoder} &  \multicolumn{2}{c|}{Planning Task} & \multirow{2}{*}{\makecell{Refine\\-ment}} & \multirow{2}{*}{PDMS} \\
        \cline{2-4}
        & VGGT & Target & Path & &\\ 
        \midrule
        1 & & & & & 85.0  \\
        2 & & \checkmark & &\checkmark & 86.0  \\
        3 & & & \checkmark &\checkmark & 86.1  \\
        4 & & \checkmark & \checkmark &\checkmark & 86.5  \\
        5 & \checkmark & & \checkmark &\checkmark & 87.2  \\
        6 & \checkmark & \checkmark & &\checkmark & 87.1  \\
        7 & \checkmark & \checkmark & \checkmark & & 87.6 \\
        8 & \checkmark & \checkmark & \checkmark &\checkmark & \textbf{87.8} \\
        \bottomrule
    \end{tabular}
    }
    \label{tab:ablation}
\end{table}

\subsection{Ablation study of scalability}
We investigate the scalability of our model by varying the backbone architecture and feature dimensions. As shown in Table~\ref{tab:scale}, increasing model capacity consistently improves performance. ResNet-101 achieves the best results (L2: 0.46 m, Collision: 0.04\%), representing a 4.2\% improvement in L2 error and 20\% reduction in collision rate compared to ResNet-50 with the same feature dimension. The results also show that feature dimension has a significant impact, with dimension 256 providing the optimal balance between performance and computational efficiency.

\begin{table}[t]
    \centering
    \setlength{\tabcolsep}{0.025\linewidth}
    \renewcommand{\arraystretch}{1.0}
    \caption{Ablation study of model scalability on nuScenes. Different backbone architectures and feature dimensions are evaluated.}
    \resizebox{0.46\textwidth}{!}{
    \begin{tabular}{c|cc|cc}
        \toprule
        ID & Backbone & Dimension & L2 (m) $\downarrow$ & Collision (\%) $\downarrow$ \\ 
        \midrule
        1 & ResNet-34 & 256 & 0.50 & 0.09 \\
        2 & ResNet-50 & 128 & 0.51 & 0.08 \\
        3 & ResNet-50 & 256 & 0.48 & 0.05 \\
        4 & ResNet-50 & 384 & 0.46 & 0.06 \\
        5 & ResNet-101 & 256 & \textbf{0.46} & \textbf{0.04} \\
        \bottomrule
    \end{tabular}
    }
    \label{tab:scale}
\end{table}


\subsection{Ablation Study on Refinement Iterations}
Table~\ref{tab:refinement} analyzes the impact of refinement iterations $K$ on model performance. The results show that $K=3$ achieves the optimal balance between performance and computational efficiency. While a single iteration provides some improvement, increasing to 3 iterations reduces collision rate by 37.5\% (0.08 → 0.05). Further iterations ($K=6$) provide diminishing returns, with collision rate slightly increasing to 0.07, suggesting potential overfitting.

\begin{table}[t]
    \centering
    \small
    \caption{Ablation study of refinement iterations $K$ on nuScenes. Different numbers of iterations in the trajectory refinement module are evaluated.}
 {
    \begin{tabular}{c|cc}
        \toprule
         Iterations & L2 (m) $\downarrow$ & Collision (\%) $\downarrow$ \\ 
        \midrule
        1 & 0.49 & 0.08 \\
        3 & \textbf{0.48} & \textbf{0.05} \\
        6 & 0.48 & 0.07\\
        \bottomrule
    \end{tabular}
    }
    \label{tab:refinement}
\end{table}

\subsection{Ablation Study on Latent Encoder Module}
We evaluate the individual contributions of VGGT and Grounded-SAM in our latent encoder module. As shown in Table~\ref{tab:latent}, VGGT alone (ID 1) provides stronger benefits than Grounded-SAM alone (ID 2), reducing L2 error by 5.8\% compared to using only Grounded-SAM. However, combining both components (ID 3) yields the best performance, demonstrating their complementary nature: VGGT provides geometric understanding while Grounded-SAM enhances semantic awareness.

\begin{table}[t]
    \centering
    \small
    \caption{Ablation study of latent encoder module components on nuScenes.}
    {\begin{tabular}{c|cc|cc}
        \toprule
        ID & VGGT& Grounded-SAM& L2 (m) $\downarrow$& Collision (\%) $\downarrow$\\
        \midrule
        1 & \checkmark &  &0.49& 0.06\\
        2 & & \checkmark  &0.52& 0.08\\
        3& \checkmark & \checkmark &\textbf{0.48}& \textbf{0.05}\\
        \bottomrule
    \end{tabular}
    }
    \label{tab:latent}
\end{table}

\begin{table}[t]
    \centering
    \small
    \caption{Ablation study of spatial path configuration on nuScenes. ``Num" denotes the number of trajectory points, and ``Inter" denotes the interval distance in meters.}
    {
    \begin{tabular}{cc|cc}
        \toprule
        \multicolumn{2}{c|}{Spatial Path Configuration} & \multirow{2}{*}{L2 (m) $\downarrow$} & \multirow{2}{*}{Collision (\%) $\downarrow$}  \\
        \cline{1-2}
        Num & Inter (m) & &  \\
        \midrule
        30& 1& 0.50& 0.12\\
        30& 2& 0.48& 0.10\\
        50& 1& 0.49& 0.09\\
        50& 2& \textbf{0.48}& \textbf{0.05}\\
        80& 1& 0.52& 0.11\\
        80& 2& 0.51& 0.12\\
        \bottomrule
    \end{tabular}
    }
    \label{tab:spatial}
\end{table}

\subsection{Ablation Study on Spatial Path Configuration}
Table~\ref{tab:spatial} investigates different spatial path configurations by varying the number of points and intervals. 
The results demonstrate that 50 points with 2 m intervals provide the optimal configuration. 
Configurations with too few points (30) or too many points (80) lead to suboptimal performance, suggesting that appropriate spatial resolution is crucial for effective path representation. 
The 2 m interval consistently outperforms 1 m interval across different point numbers, indicating that overly dense sampling may introduce noise.

\subsection{Ablation Study on Target Prediction}
We compare deterministic target point prediction with our probabilistic target region approach. As shown in Table~\ref{tab:target}, probabilistic modeling significantly reduces both the L2 error (by 9.4\%) and the collision rate (by 50\%). This demonstrates the effectiveness of our design in handling prediction uncertainty using Laplace distributions, which better capture the complexity of driving decisions. By modeling uncertainty, our approach enables the model to consider a broader range of information and make safer decisions.

\begin{table}[t]
    \centering
    \small
    \caption{Comparison between deterministic and probabilistic target modeling approaches on nuScenes.}
    {
    \begin{tabular}{c|cc}
        \toprule
        Target Modeling Method& L2 (m) $\downarrow$ & Collision (\%) $\downarrow$ \\ 
        \midrule
        Deterministic Target Point& 0.53& 0.10\\
        Probabilistic Target Region& \textbf{0.48}& \textbf{0.05}\\
        \bottomrule
    \end{tabular}
    }
    \label{tab:target}
\end{table}

\begin{figure*} [!htbp]
    \centering
    \includegraphics[width =1\linewidth]{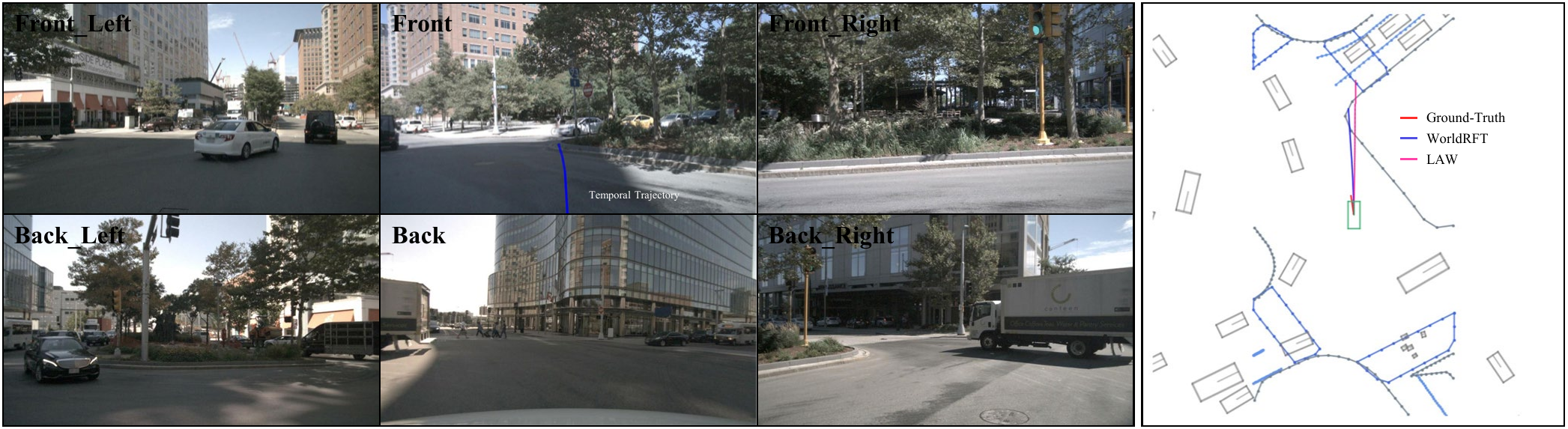}
    \caption{Visualization of failure case on nuScenes, with the input images on the left and the model output results on the right, where the red line represents the ground truth trajectory, the blue line indicates the trajectory predicted by WorldRFT, and the pink line shows the trajectory predicted by LAW \cite{li2024enhancing}.}
    \label{fig:3}
\end{figure*}

\begin{table}[t]
    \centering
    \small
    \caption{Ablation study of dataset percentage used for reinforcement learning fine-tuning on nuScenes.}
    {
    \begin{tabular}{c|cc}
        \toprule
        Percentage of Data& L2 (m) $\downarrow$ & Collision (\%) $\downarrow$ \\ 
        \midrule
        10\% & 0.47& 0.12\\
        20\% & 0.47& 0.10\\
        100\% & 0.48& \textbf{0.05}\\
        \bottomrule
    \end{tabular}
    }
    \label{tab:rl_data}
\end{table}

\subsection{Ablation Study on Dataset Percentage for RFT}
Table~\ref{tab:rl_data} investigates the impact of training data size on reinforcement fine-tuning (RFT) performance. While using only 10\% or 20\% of data achieves competitive L2 error (0.47 m), the collision rate significantly reduces when using the full dataset. This reduction in collision rate with full data demonstrates the importance of diverse scenarios for learning robust safety policies.

\begin{table}[t]
    \centering
    \small
    \caption{Comparison between supervised fine-tuning (SFT) and reinforcement fine-tuning (RFT) for safety optimization on nuScenes.}
    {
    \begin{tabular}{c|cc}
        \toprule
        Optimization Method& L2 (m) $\downarrow$ & Collision (\%) $\downarrow$ \\ 
        \midrule
        SFT & 0.48& 0.10\\
        RFT & \textbf{0.48}& \textbf{0.05}\\
        \bottomrule
    \end{tabular}
    }
    \label{tab:sft_rft}
\end{table}

\begin{table} [!htbp]
    \centering
    \small
    \caption{Ablation study of the group number of GRPO on nuScenes.}
    {
    \begin{tabular}{c|cc}
        \toprule
        Group Number & L2 (m) $\downarrow$ & Collision (\%) $\downarrow$ \\ 
        \midrule
        5 & 0.48& 0.08\\
        10 & \textbf{0.48}& \textbf{0.05}\\
        15 & 0.48& 0.07\\
        \bottomrule
    \end{tabular}
    }
    \label{tab:grpo}
\end{table}


\subsection{Safety Optimization: SFT vs. RFT}
Supervised fine-tuning (SFT) adopts the same safety-aware loss as our reinforcement learning methods but is trained using supervised learning. We compare SFT with RFT in Table~\ref{tab:sft_rft}. While both methods achieve similar L2 errors (0.48 m), RFT reduces the collision rate by 50\% (0.10 → 0.05). This significant improvement in safety performance highlights the advantage of the reinforcement learning framework: unlike SFT, which merely imitates expert trajectories, RFT leverages a collision-aware reward and GRPO to enable the model to actively learn collision avoidance.

\subsection{Ablation Study on GRPO Group Number}
Table~\ref{tab:grpo} investigates the impact of group number $G$ in GRPO. The results show that $G=10$ provides the optimal balance between exploration diversity and computational efficiency. Smaller groups ($G=5$) result in less stable advantage estimation, while larger groups ($G=15$) increase computational cost without significant performance gains.




\section{Additional Qulitative Results}

\subsection{More Visualization}
We provide additional visualization results of WorldRFT on nuScenes, as shown in Figure~\ref{fig:1} and Figure~\ref{fig:2}. On the left side is the input image, and on the right side is the corresponding schematic of the predicted results. As illustrated, WorldRFT is capable of predicting plausible and feasible trajectories, with reasonable estimates for both the target region and the spatial path. Notably, the learned target region reflects the model’s uncertainty in a physically meaningful way.

In Figure~\ref{fig:1}, which depicts a straight driving scenario, the predicted target region takes the shape of a narrow, elongated ellipse aligned primarily along the longitudinal axis. This indicates that the uncertainty is primarily attributable to ambiguity in speed prediction; although the direction of motion remains certain, the model accommodates a range of possible travel distances over the prediction horizon.

In contrast, Figure~\ref{fig:2} illustrates a scenario where the target region forms a larger, more isotropic ellipse. In this case, the center of the target region shows a degree of inaccuracy in predicting the exact future position. 
The uncertainty arises not only from speed but also from direction predictions.
However, the probabilistic nature of our approach, through the inclusion of a target region and its associated uncertainty, allows the model to consider a broader range of potential outcomes. 
This encourages more cautious decision-making, preventing the vehicle from directly heading towards the potentially inaccurate center of the target region. 

These visualization results demonstrate that our probabilistic target region, modeled via Laplace distributions, adaptively captures multimodal uncertainty in real-world driving scenarios. The shape and orientation of the predicted ellipses provide intuitive insights into the model’s confidence, enabling safer and more robust decision-making by accounting for both longitudinal and lateral prediction uncertainty. This highlights the effectiveness of our probability modeling and uncertainty handling in enhancing driving decisions and safety.

\subsection{Failure Case}

Figure~\ref{fig:3} illustrates a failure case of WorldRFT on the nuScenes, where the model produces suboptimal trajectories in response to erroneous driving directives. An analysis reveals annotation inaccuracies within the dataset—for instance, a maneuver requiring a left turn is incorrectly labeled as ``go straight"—which introduces ambiguity and hinders the generation of precise motion predictions.




\end{document}